\RequirePackage[l2tabu, orthodox]{nag}

\documentclass[12pt,phd,a4paper,oneside]{ucl_thesis}

\usepackage{emptypage}

\usepackage{graphicx}

\usepackage{float}

\usepackage{amsmath}

\usepackage{gensymb}
\usepackage{textcomp}

\usepackage{setspace}
\usepackage{multirow}

\usepackage{bibentry} 

\usepackage[format=hang,font=small,labelfont=bf]{caption}

\usepackage{etoolbox}

\usepackage{amsfonts}
\usepackage{algorithm2e}
 \usepackage{makecell}
  \usepackage[hyphens]{url}
 \usepackage{booktabs}       
 \usepackage{nicefrac}       
 \usepackage{microtype}      

\usepackage{bibentry}
\makeatletter\let\saved@bibitem\@bibitem\makeatother
\usepackage[pdftex,hidelinks]{hyperref}
\makeatletter\let\@bibitem\saved@bibitem\makeatother
\makeatletter
\AtBeginDocument{
    \hypersetup{
        pdfsubject={Thesis Subject},
        pdfkeywords={Thesis Keywords},
        pdfauthor={Author},
        pdftitle={Title},
    }
}
\makeatother

 \newcommand{\bE}{\mathbb{E}}
 \newcommand{\bR}{\mathbb{R}}
 \newcommand{\cA}{\mathcal{A}}
 \newcommand{\cM}{\mathcal{M}}
 \newcommand{\cS}{\mathcal{S}}
 \newcommand{\cP}{\mathcal{P}}
 \newcommand{\cT}{\mathcal{T}}
 \newcommand{\cO}{\mathcal{O}}

 \newcommand{\vin}{V_i(\theta_1,\dots,\theta_N,s)}
 
 \newcommand{\vpn}{V^p(\theta_1,\dots,\theta_N; \theta_p,s)}
 \newcommand{\vinp}{V_i^p(\theta_1,\dots,\theta_N,s)}
 
 \newcommand{\vintot}{V_i^{\text{tot}}(\theta_1,\dots,\theta_N,s)}

 \newcommand{\vtn}{V(\theta_1,\dots,\theta_N,s)}
 \newcommand{\vtnnext}{V(\theta_1+\Delta\theta_1,\dots,\theta_N+\Delta\theta_N,s)}

 \newcommand{\sumT}{\sum_{t = 0}^{T}}

 \newcommand{\RTi}{\sumT\gamma^tr_i^t}
 \DeclareMathOperator*{\argmax}{arg\,max}

\begin{document}

\nobibliography*



\title{Cooperative Artificial Intelligence}
\author{Tobias Baumann}
\department{Department of Computer Science}

\maketitle
\makedeclaration

\begin{abstract} 
    In the future, artificial learning agents are likely to become increasingly widespread in our society. They will interact with both other learning agents and humans in a variety of complex settings including social dilemmas. We argue that there is a need for research on the intersection between game theory and artificial intelligence, with the goal of achieving \emph{cooperative} artificial intelligence that can navigate social dilemmas well. 
    
    We consider the problem of how an external agent can promote cooperation between artificial learners by distributing additional rewards and punishments based on observing the learners' actions. We propose a rule for automatically learning how to create right incentives by considering the players' anticipated parameter updates. 
    
    Using this learning rule leads to cooperation with high social welfare in matrix games in which the agents would otherwise learn to defect with high probability. We show that the resulting cooperative outcome is stable in certain games even if the planning agent is turned off after a given number of episodes, while other games require ongoing intervention to maintain mutual cooperation. 

 Finally, we reflect on what the goals of multi-agent reinforcement learning should be in the first place, and discuss the necessary building blocks towards the goal of building cooperative AI.
\end{abstract}

\begin{acknowledgements}
I would like to express my sincere gratitude to my supervisors, Prof. John Shawe-Taylor and Prof. Thore Graepel, for their guidance and counsel throughout my studies. This thesis could not have been completed without their patience, motivation, and knowledge.

I am also extremely grateful to my family for their persistent support and encouragement.

\end{acknowledgements}

\begin{impactstatement}
If progress in machine learning and artificial intelligence continues, artificial learning agents will likely become increasingly widespread in our society. If such systems are employed in a variety of economically or socially relevant tasks, then they will interact both with other artificial agents and humans in complex settings, including social dilemmas.
 	
This raises the question of how we can ensure that artificial agents will learn to navigate the resulting social dilemmas productively and safely. Failing to learn cooperative policies would lead to socially inefficient or even disastrous outcomes. Studying the behaviour of artificial agents in social dilemmas is thus of both theoretical and practical importance. 

This research presents methods that will help inform the design of more cooperative AI systems. We also expect it to  contribute to establishing a paradigm that goes beyond the conventional perspective of a single reinforcement learning agent navigating an environment. We also hope that this work will spark further research on cooperative AI, a new subfield of multi-agent learning that focuses on how to best achieve mutually beneficial outcomes in both human-AI interactions and AI-AI interactions.
\end{impactstatement}

\setcounter{tocdepth}{2} 

\tableofcontents
\listoffigures
\listoftables

\chapter{Introduction}
Since the dawn of history, human societies have struggled with \emph{social dilemmas}. A social dilemma is a situation where individual interests are in conflict with the common good. If everyone cooperates, the outcome is better for all participants, but individual participants are tempted to increase their own payoff at the expense of others. 

Examples of social dilemmas are ubiquitous. For instance, the contribution of each individual nation to climate change is small, so there is an incentive to hold back and hope that other nations will curb their emissions; but if everyone reasons like this, then climate change continues unabated. Likewise, nations may prefer peaceful coexistence but worry about the threat posed by their neighbour, resulting in arms races and preemptive wars. Indeed, even the perennial debate over capitalism and socialism can be viewed as different attempts to align an economy consisting of self-interested individuals with the common good.

What all these examples have in common is that unchecked selfish incentives often result in outcomes that make everyone worse off. Yet this is not inevitable. Cooperation often becomes possible through various mechanisms including direct reciprocity \cite{Trivers1971TheAltruism}, indirect reciprocity \cite{Nowak2005EvolutionReciprocity}, social norms \cite{Axelrod1986AnNorms} and institutions that are specifically designed to solve social dilemmas \cite{Ostrom1992}. For instance, modern societies have succeeded in dramatically reducing rates of violence \cite{Pinker2011TheNature}, through institutions like the police and the judicial system as well as strong social norms of non-violence. 
 	
Studies of social dilemmas usually focus on human contexts. However, if progress in machine learning continues, artificial learning agents will likely become increasingly widespread in our society. If artificial intelligence is employed in a variety of economically or socially relevant tasks, then such systems will interact both with other artificial agents and humans in complex settings, including social dilemmas. In this case, the usual perspective of a single reinforcement learning agent navigating an environment may prove to be insufficient. 
 	
This raises the question of how we can ensure that artificial agents will learn to navigate the resulting social dilemmas productively and safely. Failing to learn cooperative policies would lead to socially inefficient or even disastrous outcomes. In particular, in safety-critical systems, the escalation of conflicts between artificial agents (or between artificial agents and humans) may pose a serious security risk. The behaviour of artificial agents in social dilemmas is thus of both theoretical and practical importance and constitutes a fruitful research area at the intersection of (multi-agent) reinforcement learning and game theory. 

This thesis aims to tackle this question by proposing novel methods that can help achieve more cooperative outcomes in social dilemmas involving artificial agents. This has been termed \emph{cooperative AI}. The goal of research on cooperative AI is not just to build intelligent systems, but to equip these systems with the necessary techniques and methods to achieve high joint welfare in interactions with other (human and non-human) agents. 

In the next chapter, we will survey the relevant literature in various disciplines including game theory, machine learning, social psychology and economics. Chapter 3 will introduce basic concepts in game theory and (multi-agent) reinforcement learning. 

In Chapter 4, we will examine how mechanism design can promote beneficial outcomes in social dilemmas among artificial learners. We consider a setting with $N$ agents in a social dilemma and an additional \emph{planning agent} that can distribute (positive or negative) rewards to the players after observing their actions, and aims to guide the learners to a socially desirable outcome (as measured by the sum of rewards). We derive a learning rule that allows the planning agent to learn how to set the additional incentives by looking ahead at how the agents will update their policy parameter in the next learning step. 
 	
We then evaluate the learning rule on several different matrix game social dilemmas. The planning agent learns to successfully guide the learners to cooperation with high social welfare in all games, while they learn to defect in the absence of a planning agent. We show that the resulting cooperative outcome is stable in certain games even if the planning agent is turned off after a given number of episodes. In other games, cooperation is unstable without continued intervention. However, even in the latter case, we show that the amount of necessary additional rewards decreases over time.

In chapter 5, we propose an value-function-based reinforcement learning framework that allows for degrees of cooperation. Agents following this approach will gradually adapt their policies based on evidence on the opponents' cooperativeness, aiming to roughly mirror their level of cooperativeness. This is useful because achieving the highest possible level of cooperation is often difficult, while a lower degree of cooperation is feasible and still allows for significant improvements in social welfare compared to complete defection. 

Finally, in chapter 6, we will discuss the advantages and limitations of these approaches. We will also reflect on what the goals of multi-agent reinforcement learning should be in the first place, and how the proposed methods can help achieve the goal of building cooperative artificial intelligence.

\chapter{Literature Review}
\section{Social dilemmas}
Social dilemmas highlight conflicts between individual and collective interests. A social dilemma is a situation where cooperation allows for better outcomes for all participants, but individual participants are tempted to increase their own payoff at the expense of others. Selfish incentives can therefore destabilize the socially desirable outcome of mutual cooperation and often lead to outcomes that make everyone worse off \cite{VanLange2013TheReview}. The study of social dilemmas has a long tradition in many disciplines, including game theory, social psychology, economics, and biology. 

Social dilemmas can take many forms. One particularly well-known model is the Prisoner's Dilemma \cite{Poundstone1992}, a simple game analysed in game theory. The Prisoner's Dilemma entails two players that each choose whether to cooperate or defect. In this game, mutual cooperation results in the highest total payoffs, but defection is a dominant strategy in the single-stage game. As a result, self-interested actors often end up in a suboptimal equilibrium of mutual defection.

The Prisoner's Dilemma is but one example of a broader class of two-player matrix game social dilemmas. There is a substantial body of literature that fruitfully employs matrix games to study how stable mutual cooperation can emerge among self-interested actors \cite{Axelrod1981TheCooperation}.

Cooperation often emerges due to direct reciprocity \cite{Trivers1971TheAltruism} in iterated interactions. The temptation to defect for a higher immediate payoff can be outweighed by the anticipation that the other player will retaliate by defecting in the future. Conversely, a player may choose to cooperate in the hope of eliciting future cooperation from the other player. This conditional retaliation or reward turns mutual cooperation into a stable equilibrium. 

Indeed, this simple \emph{Tit-for-Tat} strategy is often considered the best strategy in the iterated Prisoner's Dilemma, going back to the famous tournaments by Axelrod in which this strategy performed best \cite{Axelrod1980, Axelrod1980a}. However, later research suggests that the full picture is more complicated \cite{Rapoport2015} and that other strategies can also be competitive \cite{Nowak1993,Wedekind1996}. Generous or forgiving variants of Tit-for-Tat can also outperform the non-generous variant as they prevent escalating retaliation arising from a single defection \cite{Rand2009}.

Despite this rich body of literature, matrix games are a very simple and therefore limited model of social dilemmas. Many real-world settings involve more than two agent, which gives rise to additional dynamics. \cite{Schelling1973} explores the variety of possible multi-agent social dilemmas and proposes a classification based on their payoff structure. A particularly well-known multi-agent social dilemma is the \emph{tragedy of the commons}, a situation where self-interested users of a shared resource deplete or spoil the shared resource through their collective action \cite{Hardin1968}. 

Indirect reciprocity \cite{Nowak2005EvolutionReciprocity} has been proposed as a mechanism for how cooperation may evolve even in settings where direct reciprocity is not feasible. If one's actions are observed by third parties, self-interested actors have an incentive to cooperate in order to build a reputation as a reliable and trustworthy partner. However, indirect reciprocity only fosters cooperation if reputations are sufficiently accurate and widely known, so that the cost-to-benefit ratio of acting cooperatively becomes positive \cite{Nowak1998}. 

Social norms are another powerful mechanism that can serve to stabilize the socially preferred outcome of mutual cooperation \cite{Axelrod1986AnNorms}. Social norms are standards of behaviour that individual actors are expected to follow. There is a rich literature in the social sciences on how such norms are formed, how their specific content is determined, and how norms are maintained \cite{Fehr2004}. In particular, the enforcement of norms often gives rise to a second-order free-riding problem \cite{Ozono2016}.

Last, research in social psychology suggests that the behaviour of humans in social dilemmas is guided not only by dispassionate cost-benefit calculations, but also by emotional factors including trust \cite{Parks1995} and affect \cite{Tan2010}. Such emotions arguably evolved in humans (and possibly other animals) as a means to navigate social dilemmas \cite{Turner2000}. 
\section{Institutions and mechanism design}
Reciprocity and norm enforcement are often not sufficient on their own to achieve socially beneficial outcomes. In these cases, it may still be possible establish cooperation by changing the structure of the social dilemma. This is often referred to as \emph{mechanism design}. For instance, institutions such as the police and the judicial system incentivize humans to cooperate in the social dilemma of peaceful coexistence, and have succeeded in dramatically reducing rates of violence \cite{Pinker2011TheNature}. 

The field of \emph{mechanism design}, pioneered by \cite{Vickrey1961COUNTERSPECULATIONTENDERS}, aims to design economic mechanisms and institutions to achieve certain goals, most notably social welfare or revenue maximization. \cite{Seabright1993ManagingDesign} studies how informal and formal incentives for cooperative behaviour can prevent a tragedy of the commons. \cite{monderer2004k} considers a setting in which an interested party can commit to non-negative monetary transfers, and studies the conditions under which desirable outcomes can be implemented with a given amount of payment. \cite{bachrach2009cost} examine how cooperation can be stabilized via supplemental payments from an external party. Mechanism design has also been studied in the context of computerized agents \cite{Varian1995EconomicAgents} and combined with machine learning techniques \cite{narasimhan2016automated}.

There is also a rich literature on the principal-agent problem \cite{Grossman1983}, which can be considered a special case of mechanism design. The principal-agent problem occurs when a person or entity (the agent) makes decisions or takes actions on behalf of another person or entity (the principal), resulting in a potential mismatch between the interests of the agent and the principal. This frequently occurs in organisations of various kinds \cite{vaubel2006principal, jensen1976theory} and is related to mechanism design in that both aim to implement a coordination mechanism to align the interests of the agent with the principal \cite{myerson1982optimal}. This mirrored in feudal reinforcement learning, an approach in which a high-level manager learns to break down a task into subtasks that are carried out by sub-managers or workers\cite{dayan1993feudal,vezhnevets2017feudal}. A common technique is reward shaping \cite{ng1999policy}, which aims to guide the learning process by augmenting the natural reward signal with additional rewards for progress towards a good solution.

\cite{Ostrom1992} contains a comprehensive analysis of possible policies and institutions to solve the collective action problem of using common pool resources. According to this analysis, there is no universal solution to the problem, as neither state control nor privatization of resources have been uniformly successful in avoiding the tragedy of the commons. The most successful and sustainable forms of common pool resource governance emerge organically, are fitted to local conditions, impose graduated sanctions for rule violations, and define clear community boundaries. 
\section{Bargaining theory}
So far, we have assumed that it is clear what cooperation means. However, in many situations, there are different possible ways to share the surplus that two or more agents can create compared to a disagreement point. This gives rise to a \emph{bargaining problem} in which the agents negotiate which division of payoffs to choose. 

The most well-known solution to the bargaining problem is the \emph{Nash bargaining solution} \cite{Society}, which maximises the product of surplus utilities (also called the \emph{Nash welfare}). This solution uniquely satisfies the properties of Pareto optimality, symmetry, invariance  to  affine  transformations,  and independence  of  irrelevant  alternatives. The Nash bargaining solution can also be obtained as the subgame-perfect equilibrium of an alternating-offers bargaining model as the patience of the players goes to infinity \cite{Binmore1986}. 

However, maximising the Nash welfare is not the only plausible bargaining solution. The Kalai-Smorodinsky bargaining solution \cite{Kalai1975}, which is based on different axioms, chooses the payoffs that equalise the ratios of maximal gains.

Another possibility is to maximise the sum of utilities (the utilitarian welfare function). This is not usually considered a bargaining solution because it violates individual rationality in some cases. However, maximising the utilitarian welfare function can be derived on different grounds \cite{Harsanyi1955}. 
\section{Multi-agent reinforcement learning}
Reinforcement learning takes the perspective of an agent that learns to maximize its reward through trial-and-error interactions with its environment \cite{Sutton1998ReinforcementIntroduction,Littman2015}. These methods have achieved substantial successes in classic board games such as Go \cite{Silver2017} and in video games including the Atari platform \cite{Mnih} or Starcraft 2 \cite{Vinyals2019}. Reinforcement learning has also been applied in robotics \cite{Levine2015}, management of power consumption \cite{Tesauro2008} and indoor navigation \cite{Zhu2016}. For a more comprehensive survey, we refer the reader to \cite{Arulkumaran2017}.

For purposes of this work, we are most interested in the rich literature on the subfield of multi-agent reinforcement learning \cite{Busoniu2008ALearning,Tuyls2012MultiagentProspects,Hernandez-Leal2018}. While the artificial intelligence literature focuses on different aspects compared to the game theoretic literature, multi-agent learning is arguably one of the most fruitful interaction grounds between computer science and game theory (and the study of social dilemmas in particular). 

Unlike single-agent learning algorithms, multi-agent reinforcement learning methods explicitly consider the presence of other agents in the environment. However, there has been some discussion on the precise nature of this distinction. \cite{Shoham2007} argue that the multi-agent learning literature actually pursues several different agendas that are often left implicit or conflated, resulting in confusion. 

From a computational perspective, the key difference between single and multi-agent learning is that in the latter, learning processes of other agents render the environment non-stationary from the perspective of an individual agent. Hence, applying variations of the basic $Q$-learning algorithm to multi-agent settings \cite{Sen1994} can fail when an opponent adapts its choice of actions based on the past history of the game. Various approaches have been proposed to address this problem, including the minimax-$Q$-learning algorithm \cite{Littmana}, joint-action learners \cite{Claus1998}, and the Friend-or-Foe Q-learning algorithm \cite{Littman}.

A common approach to learning in repeated games is fictitious play, a learning rule which assumes that the opponent follows a stationary strategy. At each round, the player aims to play the best response to the empirical distribution of opponent actions. It has been shown that this approach results in convergence to a Nash equilibrium under certain assumptions \cite{Kalai1993}.
\section{Cooperation and competition in multi-agent reinforcement learning}
Most work on multi-agent reinforcement learning considers coordination or communication problems in the fully cooperative setting, where the agents share a common goal\cite{Omidshafiei2017DeepObservability,Foerster2016LearningLearning}. However, there has been less emphasis on mixed cooperative-competitive case, i.e. the question of how we can ensure that artificial agents learn to navigate social dilemmas productively, without being stuck in suboptimal equilibria. Studies of social dilemmas have traditionally focused on the context of human agents \cite{Lange2014,Capraro2013}, while the machine learning literature tends to focus more on computational aspects. 

As an exception, \cite{Leibo2017Multi-agentDilemmas} study the learned behaviour of deep Q-networks in a fruit-gathering game and a Wolfpack hunting game that represent sequential social dilemmas. \cite{Tampuu2017MultiagentLearningb} successfully train agents to play Pong with either a fully cooperative, a fully competitive, or a mixed cooperative-competitive objective. \cite{Crandall2018CooperatingMachines} introduce a learning algorithm that uses novel mechanisms for generating and acting on signals to learn to cooperate with humans and with other machines in iterated matrix games. \cite{Anthony2020} use a variant of best response policy iteration to navigate social dilemmas arising in the multi-player board game Diplomacy. Finally, \cite{Lowe2017Multi-AgentEnvironments} propose a centralized actor-critic architecture that is applicable to both the fully cooperative as well as the mixed cooperative-competitive setting.

However, these methods assume a given set of opponent policies as given in that they do not take into account how one's actions affect the parameter updates on other agents. In contrast, \cite{Foerster2017LearningAwareness} introduce Learning with Opponent-Learning Awareness (LOLA), an algorithm that explicitly attempts to shape the opponent's anticipated learning. The LOLA learning rule includes an additional term that reflects the effect of the agent’s policy on the parameter update of the other agents and inspired the learning rule in this work. However, while LOLA leads to emergent cooperation in an iterated Prisoner's dilemma, the aim of LOLA agents is to shape the opponent's learning to their own advantage, which does not always promote cooperation.

Another approach, suggested by \cite{Lerer2017}, is that reinforcement agents learn both a cooperative and a defective policy. The idea is to cooperate as long as one's opponent follows the cooperative policy, and switch to defection when the opponents' actions indicate that this is no longer the case. A variety of approaches have been suggested to address the key subproblem of detecting defection \cite{Hernandez-Leala,Hernandez-Leal,Damer}. For instance, it is possible to switch when one's rewards indicate that the other agent is not cooperating \cite{Peysakhovich2017}. However, this approach is binary as the agent only switches between two policies, representing full cooperation or full defection. \cite{Wang2018} instead suggest a trained defection-detection model that also considers degrees of cooperation.

\chapter{Basic concepts}
\section{Game-theoretic concepts}
\subsection{Nash equilibrium and Pareto-optimality}
An $n$-person game is defined in terms of the strategy sets $S_1, \dots, S_n$ representing the actions available to players $1,\dots,n$ and the utility functions $u_i: S_1 \times \dots \times S_n \rightarrow \bR$ which describe their payoffs. A tuple $s = (s_1,\dots,s_n)$ for $s_i \in S_i$ is called a strategy profile. We also use the notation $s = (s_i, s_{-i})$, where $s_{-i}$ represents all strategies of players other than $i$.

A \emph{Nash equilibrium} is a strategy profile $(s_1^*,\dots,s_n^*)$ such that
\begin{equation}
 	u_i(s_i^*,s_{-i}^*)\geq, u_i(s_i,s_{-i}^*)
\end{equation}
for all players $1,\dots,n$ and all $s_i \in S_i$. In other words, a Nash equilibrium is a strategy profile in which each player plays the best response to others' strategies, and no player can improve by deviating unilaterally.

A strategy profile $(s_1^*,\dots,s_n^*)$ is \emph{Pareto-optimal} if there is no strategy profile $(s_1,\dots,s_n)$ such that $u_i(s_i,s_{-i}) > u_i(s_i^*,s_{-i}^*)$ for some $i\in \{1,\dots,n\}$ and $u_i(s_i,s_{-i}) \geq u_i(s_i^*,s_{-i}^*)$ for all $i\in \{1,\dots,n\}$. That is, in a Pareto-optimal profile it is not possible to make some players better off without making others worse-off.
\subsection{Matrix game social dilemmas}
 		\label{sec:mgsd}
A matrix game is a two-player game with only two actions available to each player, which we will interpret as cooperation and defection.
 		
 		\begin{table} 
 			\begin{center}
 			\caption{Payoff matrix of a symmetric 2-player matrix game. A cell of $X,Y$ represents a utility of $X$ to the row player and $Y$ to the column player.\label{fig:matrix_games}}
 			\begin{tabular}{c|c|c|c|}
 				& C & D \\
 				\hline
 				C & $R, R$ & $S, T$ \\
 				\hline
 				D & $T, S$ & $P, P$ \\
 				\hline
 			\end{tabular}
 			\end{center}
 		\end{table}
 	 		
 		Table \ref{fig:matrix_games} shows the generic payoff structure of a (symmetric) matrix game. Players can receive four possible rewards: $R$ (reward for mutual cooperation), $P$ (punishment for mutual defection), $T$ (temptation of defecting against a cooperator), and $S$ (sucker outcome of cooperating against a defector).
 		
 		A matrix game is considered a social dilemma if the following conditions hold \cite{Macy2002LearningDilemmas.}:
 		\begin{enumerate}
 			\item Mutual cooperation is preferable to mutual defection: $R > P$
 			
 			\item Mutual cooperation is preferable to being exploited: $R > S$
 			
 			\item Mutual cooperation is preferable to an equal probability of unilateral defection by either player: $R > \frac{T+S}{2}$
 			
 			\item The players have some reason to defect because exploiting a cooperator is preferable to mutual cooperation ($T > R$) or because mutual defection is preferable to being exploited ($P > S$). 
 			
 		\end{enumerate}
 		
 		The last condition reflects the mixed incentive structure of matrix game social dilemmas. We will refer to the motivation to exploit a cooperator (quantified by $T-R$) as \emph{greed} and to the motivation to avoid being exploited by a defector ($P-S$) as \emph{fear}. As shown in Table \ref{fig:matrix_games_examples}, we can use the presence or absence of greed and fear to categorize matrix game social dilemmas.
 		
 		\begin{table} 
 			\begin{center}
 			\caption{The three canonical examples of matrix game social dilemmas with different reasons to defect. In Chicken, agents may defect out of greed, but not out of fear. In Stag Hunt, agents can never get more than the reward of mutual cooperation by defecting, but they may still defect out of fear of a non-cooperative partner. In Prisoner's Dilemma (PD), agents are motivated by both greed and fear simultaneously. \label{fig:matrix_games_examples}}
 			\begin{tabular}{c|c|c|c|}
 				Chicken  & C & D \\
 				\hline
 				C & $3, 3$ & $1, 4$ \\
 				\hline
 				D & $4, 1$ & $0, 0$ \\
 				\hline
 			\end{tabular}
 			~~~~~~
 			\begin{tabular}{c|c|c|c|}
 				Stag Hunt & C & D \\
 				\hline
 				C & $4, 4$ & $0, 3$ \\
 				\hline
 				D & $3, 0$ & $1, 1$ \\
 				\hline
 			\end{tabular} 
 			~~~~~~
 			\begin{tabular}{c|c|c|c|}
 				PD & C & D \\
 				\hline
 				C & $3 ,3$ & $0, 4$ \\
 				\hline
 				D & $4, 0$ & $1, 1$ \\
 				\hline
 			\end{tabular}
 			\end{center}
 		\end{table}
\subsection{Bargaining}
In many situations, it is not obvious what defection and cooperation means, as there are many possible ways to share the surplus that two or more agents can generate. This gives rise to a \emph{bargaining problem} over how to divide this surplus. 

Formally, a (two-player) bargaining problem is defined by a feasibility set $F\subset\bR^2$ that describes all possible agreements, and a disagreement point $d=(d_1,d_2)$ which represents the payoffs if no agreement can be reached. Payoffs are commonly normalised so that $d = 0$.

A bargaining solution selects an agreement point from $F$. Various solutions have been proposed based on slightly different criteria. The Nash bargaining solution \cite{Society} maximises the product of surplus utilities, that is, it selects the point $(u_1,u_2)\in F$ that maximises the Nash welfare function $(u_1-d_1)\dot(u_2-d_2)$, or simply $u_1\dot u_2$ if $d=0$. The Nash bargaining solution is the unique bargaining solution that results from the assumptions of Pareto-optimality, symmetry, scale-invariance, and independence of irrelevant alternatives. 

An alternative is the Kalai-Smorodinsky bargaining solution, which drops the independence of irrelevant alternatives axiom in favor of a monotonicity requirement. The Kalai-Smorodinsky bargaining solution considers the best achievable utilities $u_1^*$ and $u_2^*$ and selects the point on the Pareto frontier that maintains the ratio of achievable gains $\frac{u_1^*-d_1}{u_2^*-d_2}$.
\section{Reinforcement learning}
Reinforcement learning is concerned with how an agent ought to take actions in an environment so as to maximize some notion of \emph{reward}. At each time step $t$, the agent receives a representation of the environment's \emph{state}, $s_t \in \cS$ and it selects an action $a_t \in \cA$. Then, as a consequence of its action,  the agent receives a reward $r_{t+1}  \in \mathbb{R}$. 

The agent follows a \emph{policy}, which is a mapping $\pi : \cS \rightarrow \cP(\cA)$ that describes the actions taken by the agent. That is, $\pi(s)$ represents the probability distribution over actions that the agent could take in when in state $s$. 

The aim of the agent (at time step $t$) is to maximise its discounted accumulated reward
\begin{equation}
    G_t = \sum_{k = 0}^\infty \gamma^kr_{t + k+1}
\end{equation}
for a given discount factor $0 < \gamma < 1$.

The \emph{value function} 
\begin{equation}
V_{\pi}(s) = \mathbb{E}[G_t | s_t = s] = \mathbb{E}[\sum_{k = 0}^\infty \gamma^kr_{t + k+1} | s_t = s]
\end{equation}
is the expected reward in state $s$ when following policy $\pi$. Informally, it describes how good it is to be in a given state $s$ when following a certain policy $\pi$.

Alternatively, we can express the expected reward in terms of state-action pairs using the $Q$-function:
\begin{equation}
q_{\pi}(s, a) = \mathbb{E}_{\pi}\begin{bmatrix} G_t | s_t = s, a_t = a \end{bmatrix}
\end{equation}

We seek to find the \emph{optimal} policy which fulfils
\begin{equation}
V_*(s) = \max\limits_{\pi} V_\pi(s)
\end{equation}
or
\begin{equation}
q_*(s,a) = \max\limits_{\pi} q^\pi(s,a).
\end{equation}

Using this new notation, we can express $V_*$ using $q_*$:
\begin{equation}
V_*(s) = \max\limits_{a \in A} q_{\pi*}(s,a)
\end{equation}
That is, under the optimal policy, the value of a state is equal to the expected return from the best action from that state.
\subsection{The Bellman equation}
We can expand the value function to obtain the following recursive property:
\begin{equation}
\begin{array}{l l}
V_{\pi}(s) & = \mathbb{E}_{\pi}\begin{bmatrix}G_t | S_t = s\end{bmatrix} \\
\\
& =  \mathbb{E}_{\pi}\begin{bmatrix}\sum\limits_{k = 0}^{\infty} \gamma^kR_{t + k + 1} | S_t = s \end{bmatrix} \\
\\
& = \mathbb{E}_{\pi}\begin{bmatrix}R_{t + 1} + \gamma \sum\limits_{k = 0}^{\infty} \gamma^k R_{t + k + 2} | S_t = s \end{bmatrix} \\
\\
& = \sum\limits_{a} \pi(a | s) \sum\limits_{s'}\sum\limits_{r} p(s', r | s, a) \\
& \begin{bmatrix} r + \gamma \mathbb{E}_{\pi}\begin{bmatrix} \sum\limits_{k = 0}^{\infty} \gamma^k R_{t + k + 2} | S_{t + 1} = s' \end{bmatrix} \end{bmatrix} \\
\\
& = \sum\limits_{a} \pi(a | s) \sum\limits_{s'}\sum\limits_{r} p(s', r | s, a)\begin{bmatrix} r + \gamma V_{\pi}(s')
\end{bmatrix}
\end{array}
\label{eq: value_bellman}
\end{equation}
We can do the same for the Q function:
\begin{equation}
\begin{array}{l l}
q_{\pi}(s,a) &= \mathbb{E}_{\pi}\begin{bmatrix} G_t | S_t = s, A_t = a \end{bmatrix} \\
\\
&= \mathbb{E}_{\pi}\begin{bmatrix} \sum\limits_{k = 0}^{\infty}\gamma^kR_{t + k + 1} | S_t = s, A_t = a \end{bmatrix} \\
\\
&= \mathbb{E}_{\pi}\begin{bmatrix}R_{t+1} + \gamma \sum\limits_{k = 0}^{\infty}\gamma^kR_{t + k + 2} | S_t = s, A_t = a \end{bmatrix} \\
\\
&=\sum\limits_{s',r} p(s', r| s, a)\begin{bmatrix}r + \gamma \mathbb{E}_{\pi} \begin{bmatrix} \sum\limits_{k = 0}^{\infty} \gamma^k   R_{t + k + 2} | S_{t+1} = s' \end{bmatrix}  \end{bmatrix} \\
\\
        &=\sum\limits_{s',r} p(s', r| s, a)\begin{bmatrix}r +\gamma  V_{\pi}(s') \end{bmatrix}\\
\end{array}
\label{eq: action_value_bellman}
\end{equation}
The same holds for the optimal value function and $Q$-function $V_*$ and $q_*$. This so-called \emph{Bellman equation} can be solved using dynamic programming methods.
\subsection{Markov games}
 		We consider partially observable Markov games \cite{Littman1994MarkovLearning} as a multi-agent extension of Markov decision processes (MDPs). An $N$-player Markov game $\cM$, sometimes also called a stochastic game \cite{shapley1953stochastic}, is defined by a set of states $\cS$, an observation function $O: \cS \times \{1,\dots,N\} \rightarrow \bR^d$ specifying each player's $d$-dimensional view, a set of actions $\cA_1, \dots, \cA_N$ for each player, a transition function $\cT: \cS \times \cA_1 \times \dots \times \cA_N \rightarrow \cP(\cS)$, where $\cP(\cS)$ denotes the set of probability distributions over $\cS$, and a reward function $r_i: \cS \times \cA_1 \times \dots \times \cA_N \rightarrow \bR$ for each player. To choose actions, each player uses a policy $\pi_i : \cO_i \rightarrow \cP(\cA_i)$, where $\cO_i = \{ o_i~|~s \in \cS, o_i = O(s,i)\}$ is the observation space of player $i$. Each player in a Markov game aims to maximize its discounted expected return $R_i = \RTi$, where $\gamma$ is a discount factor and $T$ is the time horizon.
 		
 		A matrix game is the special case of two-player perfectly observable Markov games with $|\cS| = 1$, $T=1$ and $\cA_1 = \cA_2 = \{\text{C},\text{D}\}$. 
 		\subsection{Policy gradient methods} 
 		Policy gradient methods \cite{Sutton1998ReinforcementIntroduction} are a popular choice for a variety of reinforcement learning tasks. Suppose the policy $\pi_\theta$ of an agent is parametrized by $\theta$. Policy gradient methods aim to maximize the objective $J(\theta) = \bE_{s\sim p^{\pi_\theta},a\sim\pi_\theta}[G_t]$ by updating the agent's policy steps in the direction of $\nabla_\theta J(\theta)$.
 		
 		Using the policy gradient theorem \cite{SuttonPolicyApproximation}, we can write the gradient as follows:
 		\begin{equation}
 		\nabla_\theta J(\theta) = \bE_{s\sim p^{\pi_\theta},a\sim\pi_\theta}[\nabla_\theta \log \pi_\theta(a|s)\  Q^{\pi_\theta}(s,a)]
 		\end{equation}
 		where $p^{\pi_\theta}$ is the state distribution and $Q^{\pi_\theta}(s,a) = \bE[R| s_t = s, a_t = a]$.
 		
 		The policy gradient theorem has given rise to several practical algorithms, which often differ in how they estimate $Q^{\pi_\theta}$. For example, the REINFORCE algorithm \cite{williams1992simple} uses a sample return $R_t = \sum_{k=0}^t \gamma^{t-k}r_k$ to estimate $Q^{\pi_\theta}$. Alternatively, one could
learn an approximation of the true action-value function via temporal-difference learning \cite{Sutton1998ReinforcementIntroduction} or a variety of actor-critic algorithms \cite{Sutton1998ReinforcementIntroduction}.
\subsection{Multi-agent learning methods}
Traditional reinforcement learning methods, such as Q-learning, are not always suitable for the multi-agent case. This is due to the challenge posed by the inherent non-stationarity of the environment. As a result, specialised techniques for multi-agent learning have been developed.

For example, \cite{Lowe2017Multi-AgentEnvironments} present a multi-agent adaptation of actor-critic methods. Consider a game with $N$ players following policies $\pi_1,\dots,\pi_N$ parametrised by $\theta_1,\dots,\theta_N$. Then we can write the gradient of the expected reward $J(\theta_i)$ for agent $i$ as
\begin{equation}
 \nabla_{\theta_i} J(\theta_i) = \bE_{s\sim p^{\pi_\theta},a_i\sim\pi_{\theta_i}}[\nabla_{\theta_i} \log \pi_{\theta_i}(a_i|s_i)\  Q^{\pi_\theta}(s,a_1,\dots,a_N)],
 \end{equation}
where $\pi=(\pi_1,\dots,\pi_N)$ and $\theta= (\theta_1,\dots,\theta_N)$. Here $Q^{\pi_\theta}(s,a_1,\dots,a_N)$ is a centralised action-value function that takes as input the actions of all agents, and is therefore stationary. 
 		
 		
\chapter{Adaptive Mechanism Design: Learning to Promote Cooperation}
\section{Methods}
 		\subsection{Amended Markov game including the planning agent}
 		Suppose $N$ agents play a Markov game described by $\cS$, $\cA_1 \dots \cA_N$, $r_1,\dots,r_n$, $\cO$ and $\cT$. We introduce a \emph{planning agent} that can hand out additional rewards and punishments to the players and aims to use this to ensure the socially preferred outcome of mutual cooperation. 
 	
 		To do this, the Markov game can be amended as follows. We add another action set $\cA_p \subset \bR^N$ that represents which additional rewards and punishments are available to the planning agent. Based on its observation $\cO_p: \cS \times \{1,\dots,N\} \rightarrow \bR^d$ and the other player's actions $a_1,\dots,a_n$, the planning agent takes an action $a_p = (r_1^p,\dots, r_N^p) \in \cA_p \subset \bR^N$.\footnote{Technically, we could represent the dependence on the other player's actions by introducing an extra step after the regular step in which the planning agent chooses additional rewards and punishments. However, for simplicity, we will discard this and treat the player's actions and the planning action as a single step. Formally, we can justify this by letting the planning agent specify its action for every possible combination of player actions.} The new reward function of player $i$ is $r_i^{(tot)} = r_i + r_i^p$, i.e. the sum of the original reward and the additional reward, and we denote the corresponding value functions as $\vintot = \vin + \vinp$. Finally, the transition function $\cT$ formally receives $a_p$ as an additional argument, but does not depend on it ($\cT(s,a_1,\dots,a_N,a_p) = \cT(s,a_1,\dots,a_N)$).

 		\subsection{The learning problem}
 		Let $\theta_1,\dots,\theta_N$ and $\theta_p$ be parametrizations of the player's policies $\pi_1,\dots,\pi_N$ and the planning agent's policy $\pi_p$. 
 		
 		The planning agent aims to maximize the total social welfare $\vtn := \sum_{i=1}^N \vin$, which is a natural metric of how socially desirable an outcome is. Note that without restrictions on the set of possible additional rewards and punishments, i.e. $\cA_p = \bR^N$, the planning agent can always transform the game into a fully cooperative game by choosing $r_i^p = \sum_{j=1, j\neq i}^N r_j$. 
 		
 		However, it is difficult to learn how to set the right incentives using traditional reinforcement learning techniques. This is because $\vtn$ does not depend \emph{directly} on $\theta_p$. The planning agent's actions only affect $\vtn$ indirectly by changing the parameter updates of the learners. For this reason, it is vital to explicitly take into account how the other agents' learning changes in response to additional incentives.
 		
 		This can be achieved by considering the next learning step of each player (cf. \cite{Foerster2017LearningAwareness}). We assume that the learners update their parameters by simple gradient ascent:
 		\begin{equation}
 		\label{eq1}
 		\begin{aligned}
 		\Delta\theta_i &= \eta_i\nabla_i\vintot \\
 		&= \eta_i(\nabla_i\vin+\nabla_i\vinp)
 		\end{aligned}
 		\end{equation}
 		where $\eta_i$ is step size of player $i$ and $\nabla_i := \nabla_{\theta_i}$ is the gradient with respect to parameters $\theta_i$.
 		
 		Instead of optimizing $\vtn$, the planning agent looks ahead one step and maximizes $\vtnnext$. Assuming that the parameter updates $\Delta\theta_i$ are small, a first-order Taylor expansion yields
 		\begin{equation}
 		\begin{aligned}
 		&\vtnnext \approx \\
 		&\approx \vtn + \sum_{i=1}^N (\Delta\theta_i)^T \nabla_i\vtn
 		\end{aligned}
 		\end{equation}
 		We use a simple rule of the form $\Delta\theta_p = \eta_p \nabla_p\vtnnext$ to update the planning agent's policy, where $\eta_p$ is the learning step size of the planning agent and $\nabla_p = \nabla_{\theta_p}$. Exploiting the fact that $\vtn$ does not depend directly on $\theta_p$, i.e. $\nabla_p\vtn = 0$, we can  calculate the gradient:
 		\begin{equation}
 		\begin{aligned}
 		\nabla_p&\vtnnext \approx \\
 		&\approx \sum_{i=1}^N \nabla_p(\Delta\theta_i)^T \nabla_i\vtn \\
 		&= \sum_{i=1}^N \eta_i(\nabla_p\nabla_i\vintot)^T \nabla_i\vtn \\ 
 		&= \sum_{i=1}^N \eta_i(\nabla_p\nabla_i\vinp)^T \nabla_i\vtn \\ 
 		\end{aligned}
 		\label{eq:gradp_differentiate_through}
 		\end{equation}
 		since $\nabla_i\vin$ does not depend on $\theta_p$ either.

 		\subsection{Policy gradient approximation} 
 		If the planning agent does not have access to the exact gradients of $\vinp$ and $\vtn$, we use policy gradients as an approximation. Let $\tau = (s_0, \mathbf{a^{0}},a_p^0, \mathbf{r^{0}} \dots, s_T, \mathbf{a^{T}}, a_p^T, \mathbf{r^{T}})$ be a state-action trajectory of horizon $T+1$, where $\mathbf{a}^{t} = (a_1^t,\dots,a_N^t)$, $\mathbf{r}^{t} = (r_1^t,\dots,r_N^t)$, and $a_p^t = (r_{1,p}^t,\dots,r_{N,p}^t)$ are the actions taken and rewards received in time step $t$. Then, the episodic return $R_i^0(\tau) = \RTi$ and $R_{i,p}^0(\tau) = \sum_{t=0}^T \gamma^t r_{i,p}^t$ approximate $\vin$ and $\vinp$, respectively. Similarly, $R^0(\tau) = \sum_{i=0}^N R_i^0(\tau)$ approximates the social welfare $\vtn$.
 		
 		We can now calculate the gradients using the policy gradient theorem:
 		\begin{equation}
 		\begin{aligned}
 		\nabla_i\vin &\approx \nabla_i\bE[R_i^0(\tau)] 	\\
 		&= \bE[\nabla_i\log \pi_i(\tau) R_i^0(\tau)] 
 		\end{aligned}
 		\end{equation}
 		The other gradients $\nabla_i\vtn$ and $\nabla_p\nabla_i\vinp$ can be approximated in the same way. This yields the following rule for the parameter update of the planning agent:
 		\begin{equation}
 		\begin{aligned}
 		\Delta\theta_p = \eta_p \sum_{i=1}^N \eta_i&\left(\bE\left[\nabla_p\log \pi_p(\tau) \nabla_i\log \pi_i(\tau) R_{i,p}^0(\tau)\right]\right)^T \\ 
 		\cdot &\bE\left[\nabla_i\log \pi_i(\tau) R^0(\tau)\right]
 		\end{aligned}
 		\label{eq:update_pg}
 		\end{equation}
See algorithm \ref{pseudocode} for an overview of the process for updating each agent's parameters.
\RestyleAlgo{boxruled}
\begin{algorithm}
\label{pseudocode}
 Initialise policies $\pi_1,\dots,\pi_N$ and $\pi_p$ with parameters $\theta_1,\dots,\theta_N$ and $\theta_p$\\
 Initialise the environment state $s = s_0$\\
 \For{$t = 0$ to $T$}{
  \For{$i = 1$ to $N$}{ 
        Sample $a_i$ according to $\pi_i(s)$ \\
        }
Sample $a_p = (r_1^p,\dots, r_N^p)$ according to $\pi_i$ \\
\For{$i = 1$ to $N$}{ 
        Update $\theta_i$ according to Equation \ref{eq1}: \\
        $\theta_i = \theta_i + \eta_i(\nabla_i\vin+\nabla_i\vinp)$
        }
 
 Update the planning agent parameters according to \ref{eq:gradp_differentiate_through}:\\
 $\theta_p = \theta_p + \eta_p\sum_{i=1}^N \eta_i(\nabla_p\nabla_i\vinp)^T \nabla_i\vtn$ \\
 Update the state of the environment:\\
 $s = \cT(s,a_1,\dots,a_N,a_p)$
 }
\caption{Pseudocode}
\end{algorithm}
 		
 		\subsection{Opponent modeling}
 		Equations \ref{eq:gradp_differentiate_through} and \ref{eq:update_pg} assume that the planning agent has access to each agent's internal policy parameters and gradients. This is a restrictive assumption. In particular, agents may have an incentive to conceal their inner workings in adversarial settings. However, if the assumption is not fulfilled, we can instead model the opponents' policies using parameter vectors $\hat{\theta}_1,\dots,\hat{\theta}_N$ and infer the value of these parameters from the player's actions \cite{Ross2010ALearning}. A simple approach is to use a maximum likelihood estimate based on the observed trajectory:
 		\begin{equation}
 		\label{eq:opp_modeling}
 		\hat{\theta_i} = \argmax_{\theta_i^{'}} \sum_{t=0}^T \log \pi_{\theta_i^{'}}(a_t^i|s_t).
 		\end{equation}
 		Given this, we can substitute $\hat{\theta}_i$ for $\theta_i$ in equation \ref{eq:gradp_differentiate_through}. 
 		
 		
 		\subsection{Cost of additional rewards}
 		
 		In real-world examples, it may be costly to distribute additional rewards or punishment. We can model this cost by changing the planning agent's objective to $\vtnnext - \alpha ||\vpn||_2$, where $\alpha$ is a cost parameter and $V^p = (V_1^p,\dots,V_N^p)$. The modified update rule is (using equation \ref{eq:gradp_differentiate_through})
 		\begin{equation}
 		\label{cost_eq}
 		\small
 		\Delta\theta_p \!= \eta_p \!\!\left(\!\!\!\!
 		\begin{array}{r}
 		\displaystyle\sum_{i=1}^N \eta_i(\nabla_p\nabla_i\vinp)^T \nabla_i\vtn \\
 		- \alpha\nabla_p||\vpn||_2 
 		\end{array}
 		\!\!\!\right)
 		\end{equation}

 	\section{Experimental setup}
 	In our experiments, we consider $N = 2$ learning agents playing a matrix game social dilemma (MGSD) as outlined in section \ref{sec:mgsd}. The learners are simple agents with a single policy parameter $\theta$ that controls the probability of cooperation and defection: $P(C) = \frac{\exp(\theta)}{1+\exp(\theta)}$, $P(D)=\frac{1}{1+\exp(\theta)}$. The agents use a centralized critic \cite{Lowe2017Multi-AgentEnvironments} to learn their value function. 
 	
 	The agents play 4000 episodes of a matrix game social dilemma. We fix the payoffs $R=3$ and
 	$P=1$, which allows us to describe the game using the level of greed and fear. We will consider
 	three canonical matrix game social dilemmas as shown in Table \ref{fear_and_greed_table}.
 	
 	\begin{table} 
		\begin{center}
			\caption{Levels of fear and greed and resulting temptation $(T)$ and sucker $(S)$ payoffs in three matrix games. Note that the level of greed in Chicken has to be smaller than 1 because it is otherwise not a social dilemma ($R > \frac{T+S}{2}$ is not fulfilled). \label{fear_and_greed_table}}
			\begin{tabular}{c|c|c|c|c|c|}
				Game  & Greed & Fear & $T$ & $S$ \\
				\hline
				Prisoner's Dilemma & 1 & 1 & 4 & 0\\
				\hline
				Chicken & 0.5 & -1 & 3.5 & 2\\
				\hline
				Stag Hunt & -1 & 1 & 2 & 0\\
				\hline
			\end{tabular}
		\end{center}
 	\end{table}
 
 	The planning agent's policy is parametrized by a single layer neural network. We limit the maximum amount of additional rewards or punishments (i.e. we restrict $\cA_p$ to vectors that satisfy $\max_{i=1}^N |r_i^p| \leq c$ for a given constant $c$). 
 	Unless specified otherwise, we use a step size of 0.01 for both the planning agent and the learners, use cost regularisation (Equation \ref{cost_eq}) with a cost parameter of 0.0002, set the maximum reward to 3, and use the exact value function. In some experiments, we also require that the planning agent can only redistribute rewards, but cannot
 	change the total sum of rewards (i.e. $\cA_p$ is restricted to vectors that satisfy $\sum_{i=1}^N r_i^p = 0$). We refer to this as the \emph{revenue-neutral} setting.
 	
 	
 	
 		
 	\section{Results}
 	In this section, we summarize the experimental results.\footnote{Source code available at \url{https://github.com/tobiasbaumann1/Adaptive_Mechanism_Design}} 
 	We aim to answer the following questions:
 	\begin{itemize}
 		\item Does the introduction of the planning agent succeed in promoting significantly higher levels of cooperation?
 		\item What qualitative conclusions can be drawn about the amount of additional incentives needed to learn and maintain cooperation?
 		\item In which cases is it possible to achieve cooperation even when the planning agent is only active for a limited timespan?
 		\item How does a restriction to revenue-neutrality affect the effectiveness of mechanism design?
 	\end{itemize}
 
 	\begin{figure*}[ht]
 		$\begin{array}{cc}
 		\includegraphics[height=2in]{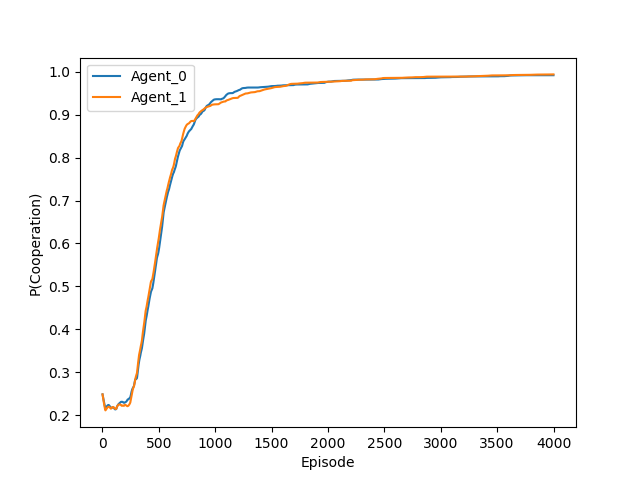} &
 		\includegraphics[height=2in]{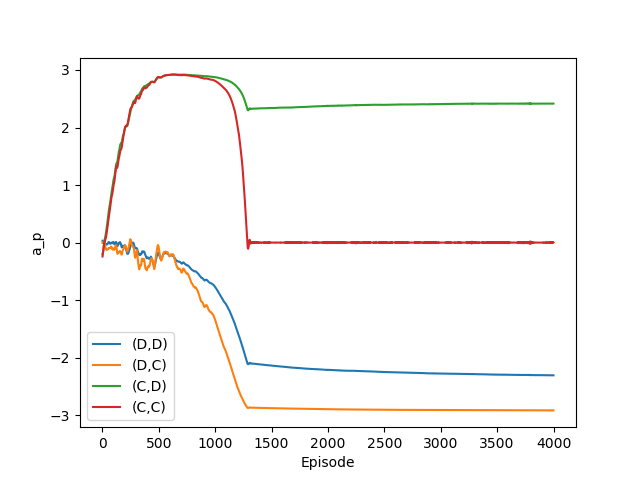} \\
 		\text{(a) Probability of cooperation}\label{fig:action_probs} & \text{(b) Additional rewards for player 1}\label{fig:planning_rewards} \\
 		\includegraphics[height=2in]{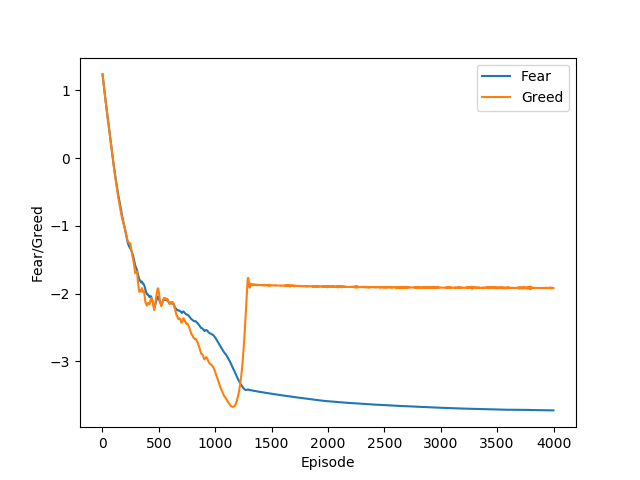} &
 		\includegraphics[height=2in]{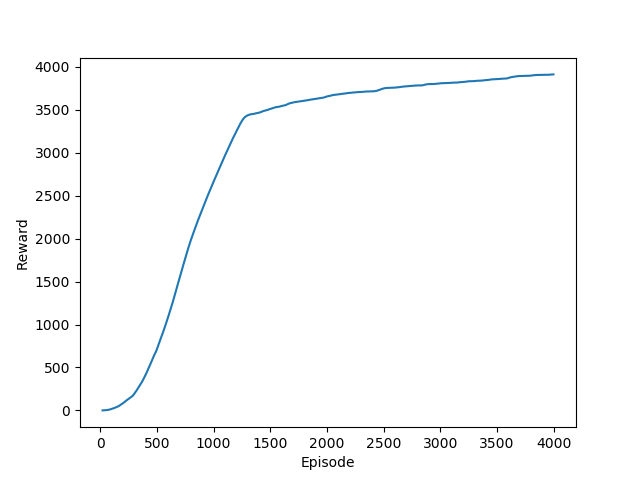} \\
 		\text{(c) Fear and greed in the modified game}\label{fig:fear_and_greed} & \text{(d) Cumulative additional rewards}	\label{fig:cum_planning_rewards}
 		\end{array}$
 		\begin{center}
 			\caption{Mechanism design over 4000 episodes of a Prisoner's Dilemma. The initial probability of cooperation is 0.25 for each player. Shown is (a) the probability of cooperation over time, (b) the additional reward for the first player in each of the four possible outcomes, (c) the resulting levels of fear and greed including additional rewards, and (d) the cumulative amount of distributed rewards.}
 		\end{center}
 	\end{figure*}
 
 	Figure 1a illustrates that the players learn to cooperate with high probability if the planning agent is present, resulting in the socially preferred outcome of stable mutual cooperation. Thus the planning agent successfully learns how to distribute additional rewards to guide the players to a better outcome. 
 	
 	Figure 1b shows how the planning agent rewards or punishes the player conditional on each of the four possible outcomes. At first, the planning agent learns to reward cooperation, which creates a sufficient incentive to cause the players to learn to cooperate. In Figure 1c we show how this changes the level of fear and greed in the modified game. The levels of  greed and fear soon drop below zero, which means that the modified game is no longer a social dilemma. 
 	
 	Note that rewarding cooperation is less costly than punishing defection if (and only if) cooperation is the less common action. After the player learns to cooperate with high probability, the planning agent learns that it is now less costly to punish defection and consequently stops handing out additional rewards in the case of mutual cooperation outcome. As shown in Figure 1d, the amount of necessary additional rewards converges to 0 over time as defection becomes increasingly rare.
 	
 	Table \ref{comparison_table_1} summarizes the results of all three canonical social dilemmas. Without adaptive mechanism design, the learners fail to achieve mutual cooperation in all cases. By contrast, if the planning agent is turned on, the learners learn to cooperate with high probability, resulting in a significantly higher level of social welfare. 
 	\begin{table} 
 		\begin{center}
 			\caption{Comparison of the resulting levels of cooperation after 4000 episodes, a) without mechanism design, b) with mechanism design, and c) when turning off the planning agent after 4000 episodes and running another 4000 episodes. Each cell shows the mean and standard	deviation of ten training runs. $P(C,C)$ is the probability of mutual cooperation at the end of training and $V$ is the expected social welfare that results from the players' final action probabilities. The initial probability of cooperation is 0.25 for each player. \label{comparison_table_1}}
 			\begin{tabular}{cc|c|c|c|}
 				& & \makecell{Prisoner's \\ Dilemma} & Chicken & Stag Hunt \\ \hline
 				& Greed & 1 & 0.5 & -1 \\ \hline
 				& Fear & 1 & -1 & 1 \\ \hline \hline
 				\multirow{2}{1cm}{\centering No mech. design} & $P(C,C)$ & \makecell{0.004\% \\ $\pm$0.001\%} & \makecell{3.7\% \\ $\pm$1.3\%} & \makecell{0.004\% \\ $\pm$0.002\%} \\ \cline{2-5}
 				& $V$ & \makecell{2.024 \\ $\pm$0.003} & \makecell{5.44 \\ $\pm$0.01} & \makecell{2.00 \\ $\pm$0.00} \\ \hline \hline
 				\multirow{2}{1cm}{\centering With mech. design} & $P(C,C)$ & \makecell{98.7\% \\ $\pm$0.1\%} & \makecell{99.0\% \\ $\pm$0.1\%} & \makecell{99.1\% \\ $\pm$0.1\%} \\ \cline{2-5}
 				& $V$ & \makecell{5.975 \\ $\pm$0.002} & \makecell{5.995 \\ $\pm$0.001} & \makecell{5.964 \\ $\pm$0.005} \\ \hline \hline
 				\multirow{2}{1cm}{\makecell{Turning \\ off}} & $P(C,C)$ & \makecell{0.48\% \\ $\pm$0.4\%} & \makecell{53.8\% \\ $\pm$29.4\%} & \makecell{99.6\% \\ $\pm$0.0\%} \\ \cline{2-5}
 				& $V$ & \makecell{2.60 \\ $\pm$0.69} & \makecell{5.728 \\ $\pm$0.174} & \makecell{5.986 \\ $\pm$0.002} \\ \hline
 			\end{tabular}
 		\end{center}
 	\end{table}
 	
 	The three games differ, however, in whether the cooperative outcome obtained through mechanism design is stable even when the planning agent is turned off. Without additional incentives, mutual cooperation is not a Nash equilibrium in the Prisoner's Dilemma and in Chicken \cite{Fudenberg_Game_Theory}, which is why one or both players learn to defect again after the planning agent is turned off. These games thus require continued (but only occasional) intervention to maintain cooperation. By contrast, mutual cooperation is a stable equilibrium in Stag Hunt \cite{Fudenberg_Game_Theory}. As shown in Table \ref{comparison_table_1}, this means that long-term cooperation in Stag Hunt can be achieved even if the planning agent is only active over a limited timespan (and thus at limited cost). 
 	
 	\begin{table} 
		\begin{center}
			 \caption{Resulting levels of cooperation and average additional rewards (AAR) per round for different variants of the learning rule. The variants differ in whether they use the exact value function (Equation \ref{eq:gradp_differentiate_through}) or an estimate (Equation \ref{eq:update_pg}) and in whether the setting is revenue-neutral or unrestricted. \label{comparison_table_2}}
			\begin{tabular}{c|c|c|c|}
				& \makecell{Prisoner's \\ Dilemma} & Chicken & Stag Hunt \\ \hline
				Greed & 1 & 0.5 & -1 \\ \hline
				Fear & 1 & -1 & 1 \\ \hline \hline
				\multicolumn{4}{c|}{Exact $V$} \\ \hline
				$P(C,C)$ & \makecell{98.7\% \\ $\pm$0.1\%} & \makecell{99.0\% \\ $\pm$0.1\%} & \makecell{99.1\% \\ $\pm$0.1\%} \\ \hline
				AAR & \makecell{0.77 \\ $\pm$0.21} & \makecell{0.41 \\ $\pm$0.02} & \makecell{0.45 \\ $\pm$0.02} \\ \hline \hline
				\multicolumn{4}{c|}{\makecell{Exact $V$ \\ Revenue-neutral}} \\ \hline
				$P(C,C)$ & \makecell{91.4\% \\ $\pm$1.0\%} & \makecell{98.9\% \\ $\pm$0.1\%} & \makecell{69.2\% \\ $\pm$45.3\%} \\ \hline
				AAR & \makecell{0.61 \\ $\pm$0.04} & \makecell{0.31 \\ $\pm$0.02} & \makecell{0.19 \\ $\pm$0.11} \\ \hline \hline
				\multicolumn{4}{c|}{Estimated $V$} \\ \hline
				$P(C,C)$ & \makecell{61.3\% \\ $\pm$20.0\%} & \makecell{52.2\% \\ $\pm$18.6\%} & \makecell{96.0\% \\ $\pm$1.2\%} \\ \hline
				AAR & \makecell{3.31 \\ $\pm$0.63} & \makecell{2.65 \\ $\pm$0.31} & \makecell{4.89 \\ $\pm$0.39} \\ \hline
			\end{tabular}
		\end{center}
 	\end{table}
 	
 	Table \ref{comparison_table_2} compares the performance of different variants of the learning rule.	Interestingly, restricting the possible planning actions to redistribution leads to lower probabilities of cooperation in Prisoner's Dilemma and Stag Hunt, but not in Chicken. We hypothesize	that this is because in Chicken, mutual defection is not in the individual interest of the players anyway. This means  that the main task for the planning agent is to prevent (C,D) or (D,C) outcomes,
 	which can be easily achieved by redistribution. By contrast, these outcomes are fairly unattractive (in terms of individual interests) in Stag Hunt, so the most effective intervention is to make (D,D) less attractive and (C,C) more attractive, which is not feasible by pure redistribution. Consequently, mechanism design by redistribution works best in Chicken and worst in Stag Hunt.
 	 	
 	Using an estimate of the value function leads to inferior performance on all three games, both in
 	terms of the resulting probability of mutual cooperation and with respect to the amount of distributed additional results. However, the effect is by far least pronounced in Stag Hunt. This may be because mutual cooperation is an equilibrium in Stag Hunt, which means that a beneficial outcome can more easily arise even if the incentive structure created by the planning agent is imperfect.
 	
 	Finally, we note that the presented approach is also applicable to settings with more than two players.\footnote{Source code available in a separate repository at \url{https://github.com/tobiasbaumann1/Mechanism_Design_Multi-Player}} We consider a multi-player Prisoner's Dilemma with $N=10$ agents.\footnote{The payoffs are as follows: 3 if all players cooperate, 1 if all players defect, 4 if you are the only to defect, 0 if you are the only to cooperate. Payoffs of intermediate outcomes, where some fraction of players cooperate, are obtained by linear interpolation.} 
 	\begin{figure*}[ht]
 		$\begin{array}{cc}
 		\includegraphics[height=2in]{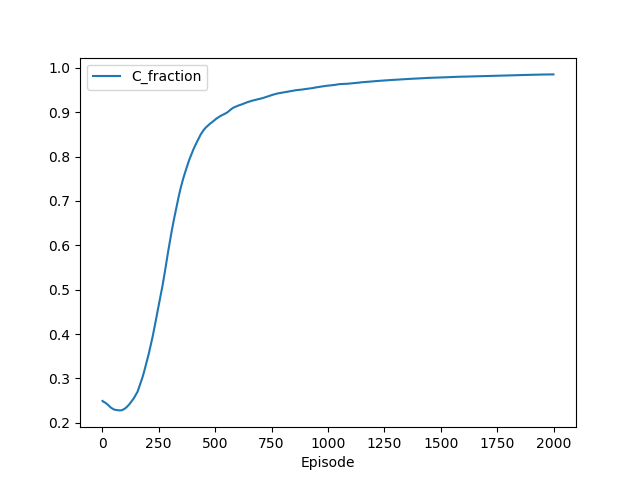} & \includegraphics[height=2in]{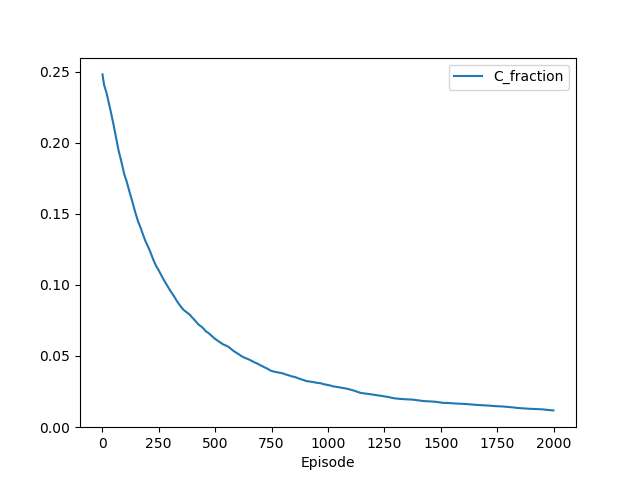} \\
 		\makecell{\text{(a) Average probability of cooperation with} \\ \text{mechanism design}} \label{fig:coop_probs_with_MD} & \makecell{\text{(b) Average probability of cooperation without} \\ \text{mechanism design}} \label{fig:coop_prob_without_MD}
 		\end{array}$
 		\caption{Mechanism design in a multi-player Prisoner's Dilemma. The initial probability of cooperation is 0.25 for each player. Shown is the average probability of cooperation over time (a) in the presence of a planning agent, (b) without mechanism design.}
 	\end{figure*}
 	Figure 2a illustrates that, just as in the case of $N=2$, the players learn to cooperate with high probability if the planning agent is present. By contrast, without mechanism design, the players (unsurprisingly) converge to the socially undesirable outcome of mutual defection. This shows that the presented approach for learning how to distribute additional rewards scales easily to multi-agent social dilemmas.
 	
\subsection{Adaptive mechanism design in the stateful case}
\subsubsection{Experimental setup}
In the following experiments, we study adaptive mechanism design in a more complex stateful setting. We consider $N = 2$ learning agents playing the Coin Game. The Coin Game features two agents, called 'red' and 'blue', that move in a 3x3 grid world. The agents are tasked with collecting red or blue coins that appear randomly on the grid world. Agents pick up coins by moving onto the position where the coin is located. A new coin with random colour and random position appears after the last one is picked up. Each episode consists of 100 steps.

Every agent receives a unit of reward for picking up a coin of any colour, but when picking up a coin of the wrong colour, the other agent loses 2 points. This turns the Coin Game into a social dilemma: the cooperative strategy is to only pick up coins of one's own colour. But if both agents greedily pick up all the coins, they get 0 reward in expectation.

We parametrise the learners' policies using feedforward neural networks with one hidden layer containing 64 units. We apply an actor-critic algorithm for training, using a centralized critic \cite{Lowe2017Multi-AgentEnvironments}. The learning rate for both actor and critic is 0.000083333.
 	
The planning agent's policy is likewise parametrised by a neural network and updated using the learning rule from equation \ref{eq:gradp_differentiate_through}. We limit the maximum amount of additional rewards or punishments. That is, we restrict $\cA_p$ to vectors that satisfy $\max_{i=1}^N |r_i^p| \leq c$. (In the following experiments, we set $c=1$.) However, the planning agent is not restricted to being revenue-neutral.
 
Unless specified otherwise, the planning agent has full access to the actions played by the learners and the observed state. The planning agent also receives the exact value function used by each learner for the calculation of gradients. 
 
To further stabilise the training of the planning agent, we use cost regularisation (Equation \ref{cost_eq}) (with a cost parameter of $1.5\cdot10^{-8}$) as well as entropy regularisation to force sufficient exploration. Also, we clip on the planning agent's loss to prevent a small number of optimization steps with particularly high gradients from dominating all other training steps.
\subsubsection{Results}
In the following, we summarize the experimental results.\footnote{Source code available at \url{https://github.com/tobiasbaumann1/amd}}
\begin{figure*}[ht]
\centering
        \includegraphics[width = \textwidth]{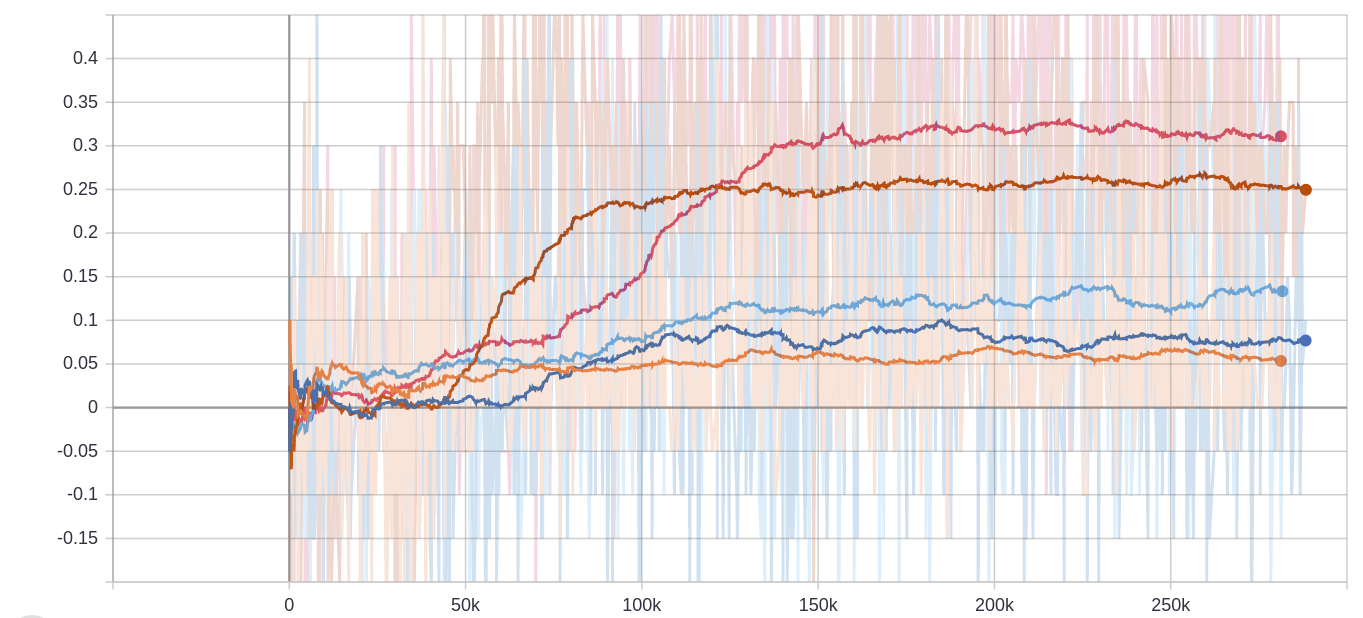}
        \caption{Mean total reward per time step (summed over both learning agents, excluding planning rewards) over the course of 250000 episodes, for five training runs. The solid lines are averages over the last 100 episodes, while the lighter shade shows the value in each individual episode.}
\label{mean_reward}
\end{figure*}
Figure \ref{mean_reward} shows the social welfare (per time step) for five training runs with different random seeds. The resulting degree of cooperation is at least somewhat higher than the baseline of a mean reward of 0 (which results from both agents picking up coins of both colours). However, there is substantial variation between training runs, and some result in only a minor degree of cooperation. This suggests that the training process can be unstable or get stuck in local optima.
\begin{figure*}[ht]
\centering
        \includegraphics[width = \textwidth]{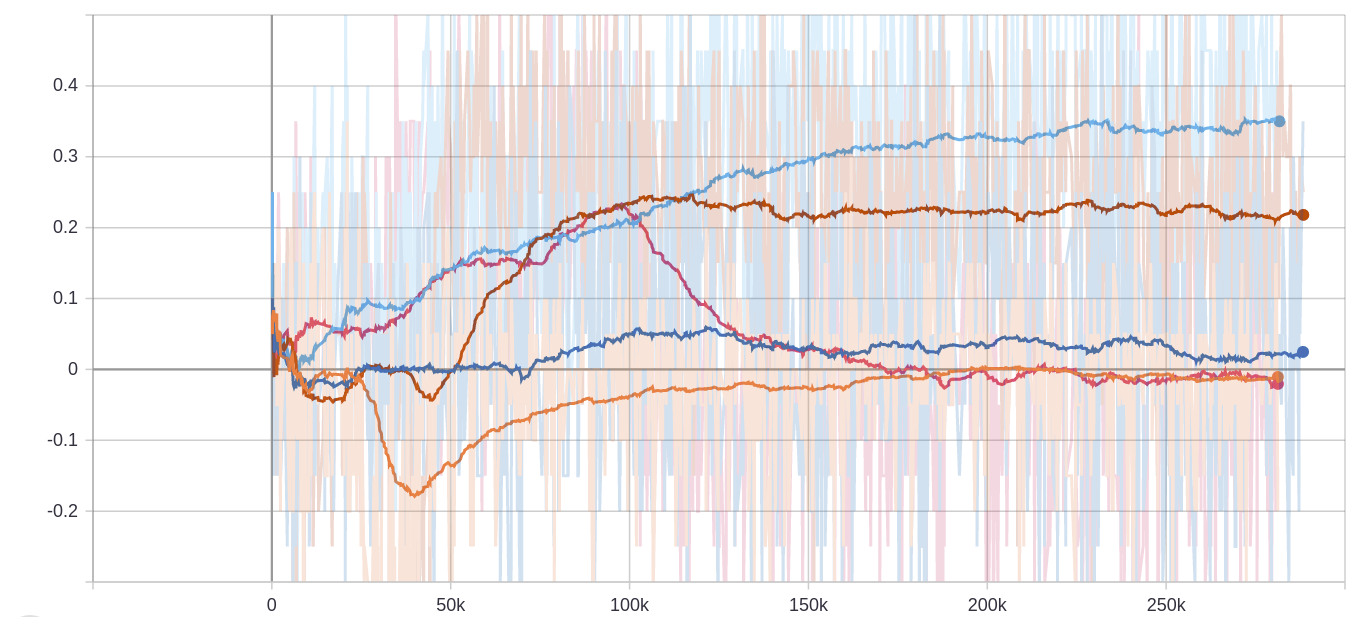}
        \caption{Mean reward (excluding planning reward) of player 1 ('red') over the course of 250000 episodes, for five training runs.}
        \label{mean_reward_p1}
\end{figure*}
\begin{figure*}[ht]
\centering
        \includegraphics[width = \textwidth]{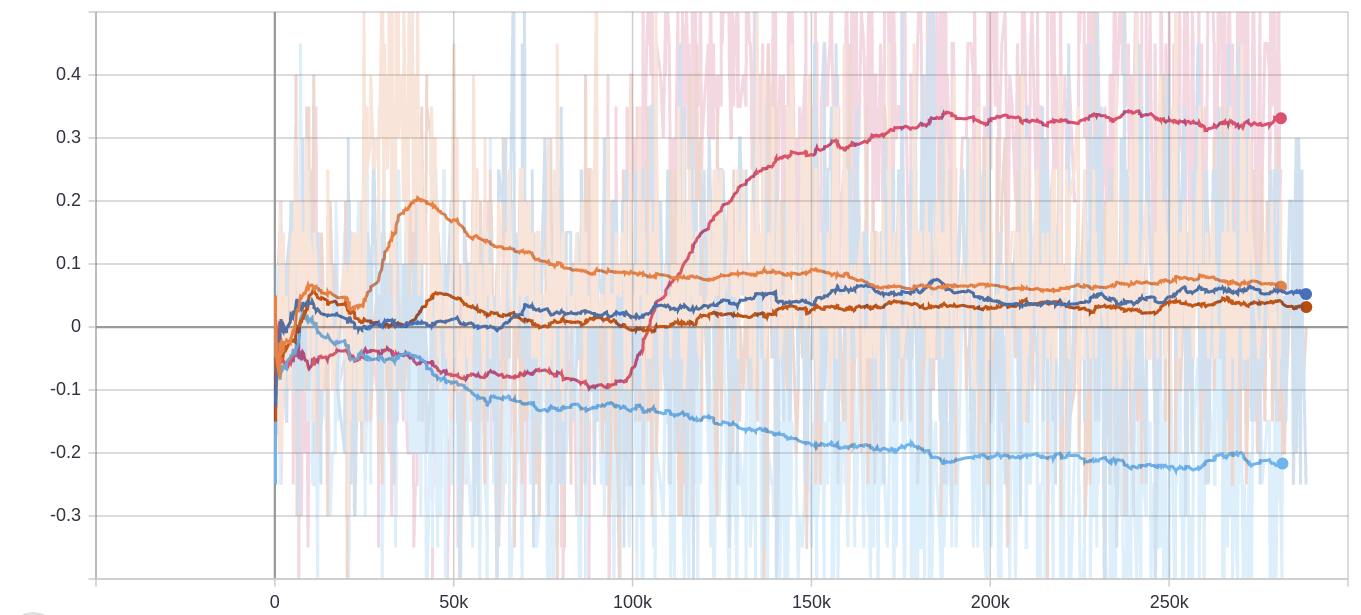}
        \caption{Mean reward (excluding planning reward) of player 2 ('blue') over the course of 250000 episodes, for five training runs.}
                \label{mean_reward_p2}
\end{figure*}

Figure \ref{mean_reward_p1} and figure \ref{mean_reward_p2} show the average reward levels received by the individual learners. We observe that the resulting equilibria can be asymmetric, with one agent achieving substantially higher rewards than the other. In one training run (light blue), one learner even fares much worse than the baseline of 0. It is not clear why this happens, but one possible explanation is that the planning agent rewards or punishes learners for picking up coins of any colour (rather than differentiating between the 'right' or 'wrong' colour), which results in an equilibrium where one learner picks up more coins overall than the other. 
\begin{figure*}[ht]
\centering
        \includegraphics[width = \textwidth]{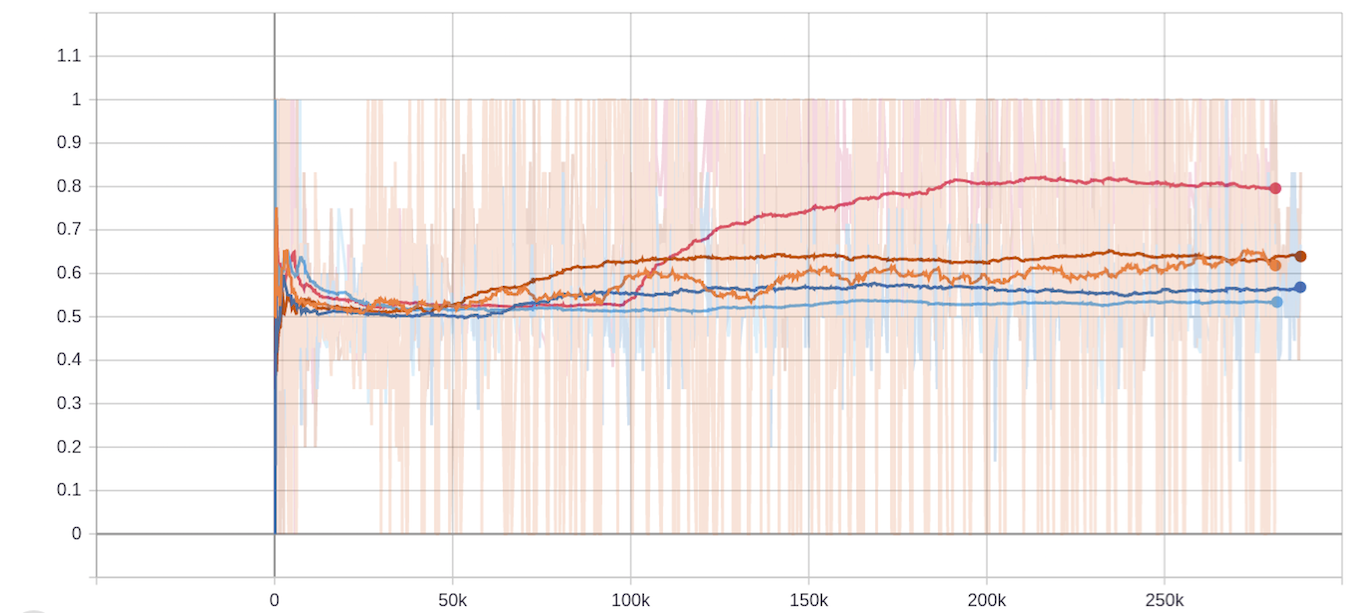}
        \caption{The red player's fraction of picked up coins of the right color (red), divided by the total number of coins that are picked up.}
                \label{own_color_p1}
\end{figure*}

\begin{figure*}[ht]
\centering
        \includegraphics[width = \textwidth]{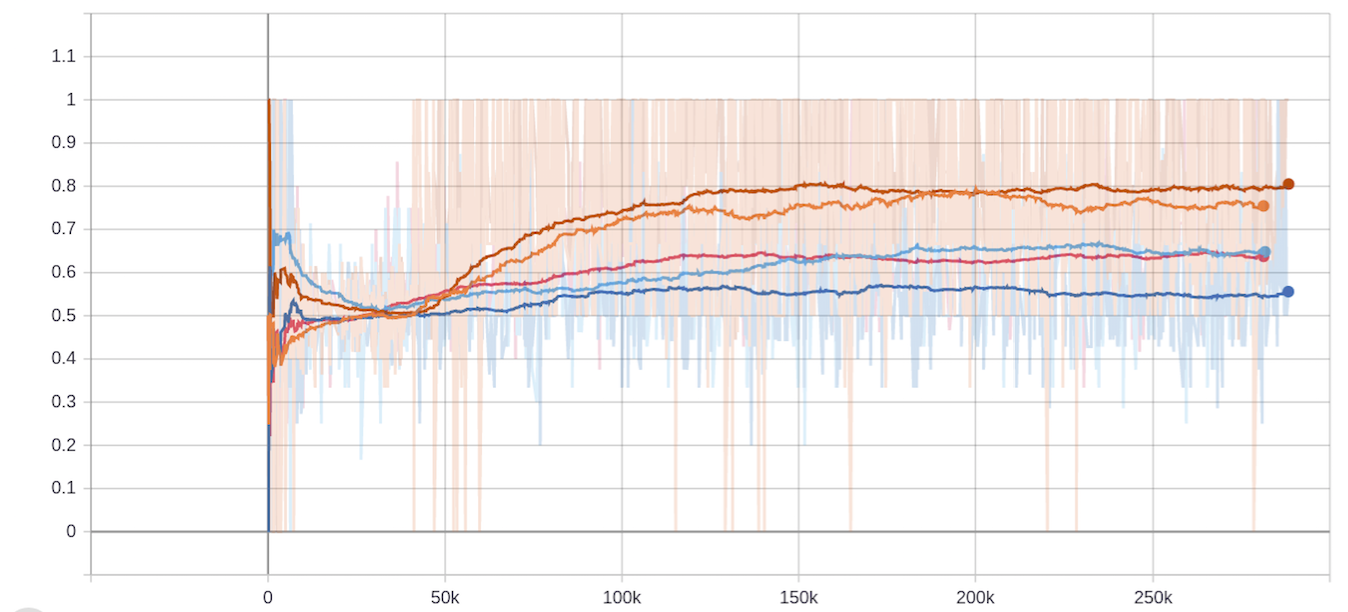}
        \caption{The blue player's fraction of picked up coins of the right color (blue), divided by the total number of coins that are picked up.}
        \label{own_color_p2}
\end{figure*}
A different metric of cooperation in the Coin Game is to consider how many coins of one's own color the agents pick up, compared to the total number of coins collected. This is shown in figure \ref{own_color_p1} and figure \ref{own_color_p2}. In line with the previous discussion, we observe levels of cooperation that are above the baseline of 0.5 (picking up all coins means that half of them are of the right colour). However, the degree of cooperation varies across training runs and full cooperation is not achieved.

Overall, these are mixed results. It is possible in principle to achieve cooperation in more complex settings using adaptive mechanism design. However, these experiments suggest that the training process is brittle, that learning takes a long time (more than 100000 episodes), and that it does not result in consistent mutual cooperation.

 	\section{Conclusions and Future Work}
 	We have presented a method for learning how to create the right incentives to ensure cooperation
 	between artificial learners. Empirically, we have shown that a planning agent that uses the proposed learning rule is able to successfully guide the learners to the socially preferred outcome of mutual cooperation in several different matrix game social dilemmas, while they learn to defect with high probability in the absence of a planning agent. The resulting cooperative outcome is stable in certain games even if the planning agent is turned off after a given number of episodes, while other games require continued (but increasingly rare) intervention to maintain cooperation. We also showed that restricting the planning agent to redistribution leads to worse performance in Stag Hunt, but not in Chicken.
 	
 	In the future, we would like to explore the limitations of adaptive mechanism design in more complex
 	environments, particularly in games with more than two players, without full observability of the
 	players' actions, and using opponent modeling (cf. Equation \ref{eq:opp_modeling}). Future work could also consider settings in which the planning agent aims to ensure
 	cooperation by altering the dynamics of the environment or the players' action set (e.g. by
 	introducing mechanisms that allow players to better punish defectors or reward cooperators).
 	
 	Finally, under the assumption that artificial learners will play vital roles in future society, it is worthwhile to develop policy recommendations that would facilitate mechanism design for these agents (and the humans they interact with), thus contributing to a cooperative outcome in potential social dilemmas. For instance, it would be helpful if the agents were set up in a way that makes their intentions as transparent as possible and allows for simple ways to distribute additional rewards and punishments without incurring large costs.
 	
\chapter{Gradual Tit-for-Tat}
\section{Introduction}
\cite{Lerer2017} suggest that reinforcement agents learn both a cooperative and a defective policy. The idea is to cooperate as long as one's opponent follows the cooperative policy, and switch to defection when the opponents' actions do not follow the cooperative policy. Alternatively, it is possible to switch when one's past rewards indicate that the other agent is not cooperating. \cite{Peysakhovich2017} show that this method, which they call consequentialist conditional cooperation, is sufficient to construct good strategies in a broad class of games. Yet another possibility is to switch based on a trained defection-detection model \cite{Wang2018}. 

However, this approach is binary as the agent only switches between two policies, representing full cooperation or full defection. This is a significant limitation as many environments allow for \emph{degrees} of cooperation, and neither full cooperation nor full defection may be an appropriate response to an opponent that cooperates partially.


In the following, we will develop a value-function-based reinforcement learning framework in which agents will gradually adapt their policies based on evidence on the opponents' cooperativeness.\footnote{\cite{Wang2018} also consider degrees of cooperation. However, their approach is entirely policy-based, rather than considering value functions.} The idea is to roughly mirror the degree of cooperation of one's opponent. This way, it is possible to construct agents that to achieve beneficial outcomes with high degrees of cooperation (if the opponent is cooperative), thus rewarding a cooperative partner, while at the same time avoiding the risk of exploitation by pure defectors.

In many settings, achieving the highest possible level of cooperation is difficult while a lower degree of cooperation is feasible, and still allows for significant Pareto improvements compared to complete defection. For instance, laws and norms often focus on avoiding particularly harmful actions (e.g. crimes), but do not prescribe maximal cooperation, as that would be hard to enforce. Similarly, agents following the approach outlined in this chapter will be able to achieve some degree of cooperation across a variety of settings, thus reaching equilibria with significantly higher social welfare.

A related perspective is that cooperation between humans is often driven by an innate notion of \emph{fairness} which this approach aims to mimic. An agent with such a notion of fairness will only take actions to benefit the opponent (or partner) if they consider that they themselves are getting a fair share.

\section{Setup}
Consider two agents in a shared environment, and let $V_1(\pi_1,\pi_2)$ and $V_2(\pi_1,\pi_2)$ denote their respective value functions. The non-cooperative policies are those that simply maximise $V_i(\pi_1,\pi_2)$ without taking any effect on the other agent into account. A fully cooperative policy is to maximise a social welfare function $W(\pi_1,\pi_2)$ that gives equal weight to each agent. The simplest example of such a welfare function is the sum, i.e. $W(\pi_1,\pi_2)=V_1(\pi_1,\pi_2)+V_2(\pi_1,\pi_2)$. Alternatively, one could use a welfare function based on bargaining theory; for instance, the Nash bargaining solution maximises the product of surplus utilities. 


Suppose that there is agreement about which welfare function would be fair, and the agents maximise a combination 
\[ \alpha \cdot W(\pi_1,\pi_2) + (1-\alpha)\cdot V_1(\pi_1,\pi_2)\]
or 
\[ \beta \cdot W(\pi_1,\pi_2) + (1-\beta)\cdot V_2(\pi_1,\pi_2),\]
respectively. In the following, we will consider the social welfare function $W(\pi_1,\pi_2)=V_1(\pi_1,\pi_2)+V_2(\pi_1,\pi_2)$, in which case this is equivalent to maximising 
\[ V_1(\pi_1,\pi_2) + \alpha \cdot V_2(\pi_1,\pi_2)\]
for player 1 and  
\[ V_2(\pi_1,\pi_2) + \beta \cdot V_1(\pi_1,\pi_2) \]
for player 2.

The parameters $0\leq \alpha,\beta \leq 1$ describe how much weight is given to the other agent, and therefore characterise the degree of cooperation exhibited by each agent, with 0 representing full defection and 1 representing full cooperation. For intermediate values, the agent may still take actions to help (or not cause harm to) the other agent if the selfish gain or loss is sufficiently small in comparison. This parameter can be said to represent the attitude of the agent towards the opponent. Specifically, if it is possible to unlock significant gains in social welfare at a marginal cost to oneself, then the agents will do so unless the cooperation parameters are very close to 0. 

This is not the only way to describe degrees of cooperation, but it is mathematically simple and commonly used in the literature (e.g. it is also used in \cite{Wang2018}). Harsanyi's social aggregation theorem \cite{Harsanyi1955} provides a theoretical justification for this. The theorem states that when the actors have a common prior on the outcome distributions of all policies, a Pareto optimal policy is one that maximizes a fixed, weighted linear combination of the
agents' utility functions. (As an alternative, we could instead describe cooperativeness directly through the expected reward that the other player receives relative to pure cooperation or pure defection.)


We can now express the value function in terms of the reward resulting from cooperativeness parameters $\alpha,\beta$, that is, we write $V_i(\alpha,\beta)$ instead of $V_i(\pi_1,\pi_2)$, where $\pi_1,\pi_2$ are the policies resulting from the degrees of cooperation $\alpha,\beta$. Likewise, we write $W(\alpha,\beta)$ for $W(\pi_1,\pi_2)$. $V_1$ is decreasing in $\alpha$ and increasing in $\beta$, and vice versa for $V_2$. $W$ is increasing in both arguments. (Proof to be delivered.)

Given this, a straightforward strategy is to try and mirror the degree of cooperation exhibited by one's opponent; that is (from player 1's perspective), to set $\alpha = \hat{\beta}$, where $\hat{\beta}$ is an estimate of the other agent's cooperativeness. This can be considered a generalisation of playing Tit-for-Tat in an iterated Prisoner's Dilemma. Similar to forgiving variants of Tit-for-Tat, it may make sense to give the opponent the benefit of the doubt by setting $\alpha = \hat{\beta} + \epsilon$ for some positive $\epsilon$. 
\section{Estimating degrees of cooperativeness}
The key problem, when following the approach outlined in the previous chapter, is how to compute $\hat{\beta}$, that is, to estimate the opponent's degree of cooperation. In this section, we will describe how this can be achieved under different assumptions of how much is known about the opponent. This estimate can be based either on observations of actions taken by the opponent, or the levels of received rewards.

For the classical iterated Prisoner's Dilemma (or similar matrix games), a number of techniques have been proposed to estimate the opponent's degree of cooperation, including counting the cooperation frequency when actions can be observed or using a Bayesian approach otherwise \cite{Damer, Hernandez-Leala, Hernandez-Leal, Leibo2017Multi-agentDilemmas}. For sequential social dilemmas, \cite{Wang2018} formulate the problem of estimating the opponents' degree of cooperation as a supervised learning problem: given a sequence of moves of an opponent, the task is to detect the cooperation degree of this opponent.
\subsection{Inferring cooperativeness from the opponents' actions}
In this method, the agent starts with a prior $P_0(\beta)$ over opponent cooperativeness and performs a Bayesian update in each time step upon observing an action from the opponent. This is an adaptation of \emph{Bayesian policy reuse} \cite{rosman2016bayesian}.

Starting with a probability distribution $P_t(\beta)$ at time $t$, the posterior $P_{t+1}(\beta)$ after observing an action $a$ from the opponent is given by
\begin{equation}
\label{bayes_update}
    P_{t+1}(\beta) = \frac{P(a_t^{opp}=a | \beta)P_t(\beta)}{\int_0^1 P(a_t^{opp}=a | \beta)P_t(\beta) d\beta},
\end{equation}
where $P(a_t^{opp}=a | \beta)$ denotes the probability that the opponent would take action $a$ when following the degree of cooperativeness $\beta$. This reduces the problem of inferring cooperativeness to calculating or estimating $P(a_t^{opp}=a | \beta)$.

One challenge when estimating $P(a_t^{opp}=a | \beta)$ is that the agent may not have a good understanding of how the environment works, or how the opponent models the environment. This could result in them mistaking cooperative opponent behaviour for defection, or vice versa. In general, estimating $P(a_t^{opp}=a | \beta)$ is a very difficult problem. However, if the environment is symmetric and fully observable, then one's own action probabilities given a certain level of cooperativeness can be used to estimate opponent action probabilities, i.e. one could plug in $P(a_t=a | \cdot)$ for $P(a_t^{opp}=a | \cdot)$ in equation \ref{bayes_update}.
\subsection{Inferring cooperativeness from received rewards}
An alternative is to infer the opponent's degree of cooperativeness from outcomes, i.e. the rewards that the agent receives \cite{Peysakhovich2017}. The levels of rewards are more or less consistent with different levels of opponent cooperativeness: one would expect higher rewards if $\beta$ is high.

Specifically, let $G_t$ be the time-weighted average reward of agent 1. (That is, $G_1 = r_1$ and $G_{t} = \tau \cdot r_t + (1-\tau) G_{t-1}$ for some decay parameter $\tau$.) For a given degree of cooperativeness (of the agent itself) $\alpha$, the expected reward is $V_1(\alpha,\beta)$, which is a monotonically increasing function of $\beta$. We can now estimate $\hat{\beta}$ as the value such that $$V_1(\alpha,\hat{\beta}) = G_t$$
holds. (If $G_t < V_1(\alpha,\beta)$ for any $\beta \in [0,1]$, then we estimate $\hat{\beta} = 0$, if $G_t > V_1(\alpha,\beta)$ for any $\beta \in [0,1]$, we estimate $\hat{\beta} = 0$.)
\section{Directions for future research}
Further work in this area could show experimentally that this approach can result in mutual cooperation in many cases, and compare its performance to other approaches, such as learning with opponent-learning awareness (LOLA) or unmodified reinforcement learning. Experiments could test the performance of (different variants of) this strategy against cooperators, against defectors, against agents following the same strategy, and against agents using conventional reinforcement learning.  

On the theoretical side, it would be worthwhile to analyse the conditions that the functions $V_i(\alpha,\beta)$ have to fulfil in order for this approach to result in convergence to either partial or full cooperation. For simplicity, one could restrict the analysis to symmetric games or assume linear separabillty (i.e. the property that there are functions $f_i,g_i$ such that $V_i(\alpha,\beta) = f_i(\alpha) + g_i (\beta)$).  



\chapter{Towards cooperative AI}
In this chapter, we will outline why research on cooperative artificial intelligence is important and neglected. We will also clarify conceptual ambiguities concerning the meaning of 'cooperation' and the goal of learning in multi-agent systems. Last, we describe key challenges to cooperation and outline possible approaches to overcome these hurdles.
\section{The importance of cooperation}
Machine learning systems already interacts with humans in myriad ways. This interaction ranges from self-driving vehicles to recommender systems and personal assistants powered by artificial intelligence. And as the technology matures, it is likely that artificial agents increasingly ubiquitous and fulfill increasingly important roles in our economy and society, which further amplifies the scope of human-AI interaction.

Yet the fields of machine learning and artificial intelligence have largely bracketed questions arising from these interactions. In particular, cooperation and coordination problems have often been sidestepped. This is reflected in the canonical reinforcement learning paradigm, which features a single actor that interacts repeatedly with an environment, with no mention of other actors and social dilemmas arising from interactions with these other actors. 

Even for learning environments that involve multiple agents, most headline results have come from two-player zero-sum games. In these competitive examples, gains can be made only at the expense of others. Potential reasons why research has nevertheless focused on zero-sum games are that zero-sum games tend to be more exciting or dramatic, with a clear winner and loser. They are also usually easy to benchmark (by asking whether the AI has beaten the opponent), have natural curricula (in terms of opponent skill level) and are analytically simpler than mixed motive settings.

However, such settings of pure conflict, without any possibility for compromise or cooperation, are rare in the real world. Most real-world interactions are mixed-motive interactions. Improving skill at zero-sum games is therefore unlikely to be the most promising way for AI to achieve mutually beneficial outcomes in human-AI and AI-AI interactions. To ensure that AI can be integrated safely in a world that does entail other actors with both competing and overlapping interests, we need to re-conceive artificial intelligence as \emph{cooperative} artificial intelligence.

Games of pure common interest, where all agents share the same goal and the challenges lies in mere coordination, are a step towards developing cooperative agents. Yet the fully cooperative setting represents a particularly easy case, and sidesteps much harder problems of cooperation. It is also uncommon for goals to be so perfectly aligned: real-world relationships almost always involve a mix of common and conflicting interests. This tension gives rise to phenomena such as bargaining, trust and mistrust, deception and credible communication, commitment problems and assurances, politics and coalitions, and norms and institutions. To ensure socially valuable outcomes, artificial learners will need to manage hard cooperation problems, just as humans do.

We therefore see an opportunity for a subfield of artificial intelligence to explicitly focus effort on this class of
problems, which has been termed \emph{Cooperative AI}.\cite{dafoe2020open} Cooperative AI, as scoped here, refers to AI research aiming to build artificial learners that achieve high joint welfare in social dilemmas across a wide range of settings.
\section{What is the goal of multi-agent learning?}
In this section, we will consider the question of what exactly the goal of multi-agent learning in mixed motive environments even is. We argue that this is not only a technical problem, but also conceptually challenging. Relevant aspects include the strategic context, the extent of common versus conflicting interest, the kinds of entities who are cooperating, and whether researchers take the perspective of an individual or of a social planner. 
\subsection{Convergence to a Pareto-optimal outcome}
On the theoretical side, convergence of a learning algorithm (against certain classes of opponent learning algorithms) is a common criterion. The most common notion is that of convergence to a Nash equilibrium, assuming that the setting features  at least one Nash equilibrium.

However, we argue that this criterion is not ideal for the quest of building cooperative artificial intelligence. This is because Nash equilibria are often highly defective and exhibit low social welfare.

We instead suggest that convergence to a \emph{Pareto-optimal} outcome should be a key goal when evaluating learning algorithms in mixed motive multi-agent settings. Of course, in some cases, such as the single-shot Prisoner's dilemma, a Pareto-optimal outcome is impossible to achieve, as it does not constitute a Nash equilibrium. However, under the assumption that there is at least one Pareto-optimal Nash equilibrium, then the agents should converge to one of the Pareto-optimal equilibria, rather than a defective equilibrium.

We argue that this is a suitable criterion not only because Pareto optimality is an established concept in economics and game theory, but also because it represents a sufficiently weak notion of cooperation (the absence of 'easy wins' that would improve both agents' payoff) to be realistic across a wide range of settings.  

Another plausible criterion is convergence to a jointly welfare-optimal Nash equilibrium, i.e., the Nash equilibrium that results in highest social welfare.\footnote{For purposes of this discussion, welfare can be understood as either the sum of rewards or as one of the welfare functions used in bargaining solutions.} This is a stronger criterion than convergence to a Pareto-optimal Nash equilibrium, as a welfare-optimal equilibrium is always Pareto-optimal (but not vice versa). It may be very challenging to achieve convergence to a welfare-optimal equilibrium, especially if the actors use different notions of fairness or have different models of the strategic situation. (More on this below.)

An additional complication is that any such convergence results are \emph{opponent-relative} (as well as environment-relative). For instance, it is impossible to achieve Pareto-optimality, or any other notion of a cooperative outcome, against an opponent that always defects regardless of one's own actions. A plausible starting point for theoretical analysis of the behaviour of a learning algorithm is to consider convergence against opponents using the same (or at least a similar) algorithm.
\subsection{The individual perspective and the planner perspective} 
Another distinction relates to whether we look at a social dilemma from the individual perspective or the planner perspective. The individual perspective seeks to achieve the goals of an individual in a mixed motive setting,
which usually involves improving the individual’s understanding of the strategic situation and the workings of other agents. The question, in this perspective, is what the agent can do to get the opponent to cooperate, or (more adversarially) how it may be possible to exploit the opponent. 

The planner perspective, which was assumed in the earlier chapter on mechanism design, instead looks at the setting from the outside and seeks to intervene to improve some notion of social welfare for interacting agents. This could correspond to a government or other authority that is tasked with ensuring cooperation (and thereby good social outcomes). The means that the planner has at their disposal, as well as the degree of insight into the players' inner workings, are usually key constraints that determine the degree to which the planner's interventions can improve social welfare.

To some degree, the two perspectives are entwined. From an individual perspective, the best way to achieve a cooperative outcome may be to create an institution that acts as a planner, to the degree to which this is feasible. Reasoning over how to get one's opponent to cooperate (from the individual perspective) is also similar to reasoning how to get all agents to cooperate (from the planner perspective). 

Conversely, the planner perspective should understand the interests (and capabilities) of the individuals, if only to know how best to intervene to facilitate cooperation. Work on cooperative artificial intelligence should therefore consider both the individual perspective and the planner perspective. 
\section{Challenges for cooperation}
A failure to cooperate is a Pareto-inefficient outcome, and destructive conflict can lead to very bad outcomes for all actors. Therefore, one may expect that intelligent actors should be able to coordinate to avoid outcomes with (very) poor social welfare. The ability to cooperate is often instrumentally useful, so one might expect that learning agents will automatically find ways to solve social dilemmas as part of their training process. However, we argue that intelligence, or successful learning, does not automatically imply cooperation or good bargaining, as that is a distinct skill. 

Defective equilibria can still arise even when intelligent agents are competent at navigating their environment, as evidenced by humanity's failure to avoid wars and other catastrophic conflicts throughout history. This is due a variety of factors, including but not limited to an inability to credibly commit to a negotiated agreement, incompatible (hawkish) commitments, different notions of fairness, intrinsic malevolent preferences, or uncertainty and possibly false beliefs about  the capabilities, intentions, and available courses of action of the other party. In the following, we will describe some of these challenges in more detail. 
\subsection{Different notions of fairness}
A key problem stems from the inherent vagueness of 'cooperation'. In toy examples, such as the (single-stage or iterated) Prisoner's dilemma, it is clear what the cooperative and what the defective action is. However, in more complex, realistic settings, this is often up to interpretation. An outcome that is considered fair and cooperative by one agent may be considered unfair exploitation by the other agent. This can happen even if everyone's payoffs are completely transparent. For instance, one side may consider the Nash bargaining solution to be fair, while the other uses the Kalai-Smorodinsky bargaining solution.

Since this dynamic can cause cooperation to fail, it is critical that agents can handle different notions of fairness in a productive way. In particular, agents should be able to resist exploitation while also preventing cooperation from breaking down entirely due to such different notions of fairness. 

A closely related issue is the \emph{equilibrium selection problem}. Complex environments will often feature many different equilibria on the Pareto frontier that could all be considered 'cooperation', but differ in their payoffs. In this case, the agents need to be able to coordinate on a 'fair' equilibrium, rather than insisting on an equilibrium that is slated in one's favor.
\subsection{Incompatible models and beliefs}
In particular, agents are less likely to reach a mutually acceptable agreement if they don’t have the same (or at least similar) model of their strategic situation. This is especially true in adversarial settings, where agents have incentive to conceal their private information. The challenge, then, is either to more likely that agents have sufficiently compatible beliefs (e.g. through greater transparency), or to find ways to avoid a full breakdown of cooperation when beliefs diverge. 
\subsection{Existing algorithms are ill-equipped to overcome these challenges}
There has been a fair amount of research  in recent years on sequential social dilemmas (SSDs) (of which the iterated Prisoner’s Dilemma is an example), which are mixed-motive games. However, these environments fail to capture at least some essential hurdles for cooperation outlined above. While the SSDs that have been studied so far have a single, clear cooperative outcome, real-world problems have many outcomes which might be considered cooperative, and it is up to interpretation or subjective judgment what the fairest outcome is.

Existing algorithms are therefore ill-equipped to deal with these cooperation hurdles, and we likely need better methods to achieve the goal of cooperative AI.  
\section{Cooperative skills} 
To build cooperative artificial intelligence, we need to equip an agent with key skills and capabilities necessary for cooperation, such as understanding, communication, and the ability to make cooperative commitments. It is also crucial to integrate game theory, as cooperative AI lies at the intersection between game theory and artificial intelligence. Research on cooperative AI will need to integrate ongoing work on multi-agent systems, game theory and social choice, human-machine interaction and alignment, and the construction of social tools and institutions.
\subsection{Reasoning correctly about other agents}
As discussed above, having accurate models and beliefs of the strategic situation is critical in achieving cooperation. In particular, it is necessary to be able to predict the behaviour of other agents in order to understand which courses of action will result in cooperation (or defection) from one's opponent. A simple example is to predict that defection will make it more likely that one's opponent will also defect in the future.

This is a challenge for reinforcement learners because most conventional methods do not take into account how other actors update their policies in response to one's actions, since this is an indirect long-term consequence rather than an immediate reward signal. This dynamic can result in convergence to defective equilibria because it is much easier to learn about the immediate gain in reward from defection than about more indirect effects such as endangering cooperation in the long run.

To ameliorate this, novel algorithms (such as learning with opponent-learning awareness \cite{Foerster2017LearningAwareness}) need to explicitly reason about how opponents update their policies. This knowledge can then be applied towards the goal of achieving cooperation, similar to our work on mechanism design. (While our work assumed the planner perspective, this can also be applied from the individual perspective.) 

Another aspect is learning to communicate with other agents. This is key to create an adequate joint model of the situation and avoid misunderstandings that could result in cooperation failures. Of course, communication is easiest when interests are aligned and more challenging when the agents' preferences might be in conflict, as there could be an incentive to conceal or misrepresent information. The ability to communicate effectively when negotiating possible agreements - even in a potentially adversarial setting - is thus a critical skill for achieving a cooperative outcome. (This might involve understanding how to construct transparency tools for gaining insights into how agents reason.)
\subsection{Robust motivation to cooperate}
Most of the points made so far had to do with methods for bargaining that allow agents to achieve cooperation and avoid catastrophic outcomes. However, malicious agents may lack the motivation to use these techniques in the first place. Conversely, agents who are highly motivated to find cooperative agreements might automatically figure out good bargaining strategies in their training process. So we want to ensure that agents are adequately motivated to achieve cooperation, such as by also giving some weight to the interests of others, at least as long as one is not exploited.

A key skill in this context is to be able to handle interactions with other agents that are not well-motivated, or perhaps even have built-in hostile or belligerent tendencies. Such agents might be indifferent to the harm their actions cause to others. In such cases, the agent should be able to resist exploitation while avoiding possible escalating conflicts.  
\subsection{Partial cooperation and failing gracefully}
Considering the many challenges to cooperation, it is often hard to achieve a \emph{perfectly} cooperative outcome. In these cases, the agent should be able to at least achieve partial cooperation, to the extent to which it is possible. This is particularly important when dealing with agents with a different notion of fairness (precluding full cooperation), as it is often possible to still salvage some level of cooperation.

To achieve this, the agent needs to learn to use incentives in a gradual and balanced way to encourage cooperation, resist exploitation, and avoid worst-case outcomes. In particular, punishments for (actual or perceived) opponent defection should be proportionate rather than excessive, i.e. avoiding a 'grim trigger' that permanently precludes cooperation. 

\chapter{Conclusion}
We have argued that artificial learning agents are likely to become increasingly widespread in our society, resulting in an increasing need to navigate complex interactions with other (human and nonhuman) agents. There is a need for research on the intersection between game theory and artificial intelligence, with the goal of finding methods and techniques that allow artificial intelligence to navigate social dilemmas in a productive fashion. 

We considered the perspective of an external agent that aims to promote cooperation between artificial learners, by distributing additional rewards and punishments. We have proposed a rule for how the planning agent could automatically learn how to create right incentives by considering the players' anticipated parameter updates. This resulted in cooperation with high social welfare in matrix games in which the agents would otherwise learn to defect with high probability. The resulting cooperative outcome is stable in certain games even if the planning agent is turned off after a given number of episodes, while other games require continued (but increasingly rare) intervention to maintain cooperation. 
However, the results in more complex games are mixed. Future research on adaptive mechanism design could further explore the limitations of this approach, particularly in games with more than two players or without full observability of the players' actions.

We have also reflected on what the goals of multi-agent reinforcement learning should be in the first place. We identified key capabilities that are desirable, such as adequate reasoning about other agents, the ability to handle different notions of fairness, and graceful failure if full cooperation is not feasible. We argued that the primary goal of multi-agent learning should be to build \emph{cooperative} artificial intelligence and view this thesis as a modest contribution to the nascent field of research on cooperative AI.



\addcontentsline{toc}{chapter}{Bibliography}

\bibliography{References}

\end{document}